\documentclass{article}
\usepackage{baichuan}
\usepackage{makecell}
\usepackage{CJKutf8}
\usepackage[utf8]{inputenc}
\usepackage[T1]{fontenc}
\usepackage{hyperref}
\usepackage{url}
\usepackage{booktabs}
\usepackage{amsfonts}
\usepackage{nicefrac}
\usepackage[ruled,linesnumbered]{algorithm2e}
\usepackage{setspace}
\SetAlgoCaptionSeparator{~}
\SetAlFnt{\small}
\DontPrintSemicolon
\usepackage{siunitx} 
\usepackage{adjustbox}
\usepackage{microtype}
\usepackage{xcolor}
\usepackage{multirow}
\usepackage{multicol}
\usepackage{amsmath}
\usepackage{graphicx}
\usepackage{amsmath}

\usepackage{subcaption}
\usepackage{utfsym}
\usepackage{svg}
\usepackage{array}
\usepackage{graphicx}
\usepackage{fancyhdr}
\usepackage{tablefootnote}
\usepackage{footnote}
\usepackage{tabularx}
\usepackage{hyperref}
\usepackage{xcolor}
\usepackage{wrapfig}

\title{Baichuan-M1: Pushing the Medical Capability \\ of Large Language Models}
\vspace{-0.7cm}
\author{
	Baichuan Inc.\thanks{The contribution of this paper are shown in \ref{sec.contributor}. Correspondent: daniel@baichuan-inc.com.}
}

\colmfinalcopy

\begin{document}
	\begin{center}
		\maketitle
	\end{center}
	\vspace{-1.0cm}
	\begin{abstract}\label{abstract}
		The current generation of large language models (LLMs) is typically designed for broad, general-purpose applications, while domain-specific LLMs, especially in vertical fields like medicine, remain relatively scarce. In particular, the development of highly efficient and practical LLMs for the medical domain is challenging due to the complexity of medical knowledge and the limited availability of high-quality data. To bridge this gap, we introduce Baichuan-M1, a series of large language models specifically optimized for medical applications. Unlike traditional approaches that simply continue pretraining on existing models or apply post-training to a general base model, Baichuan-M1 is trained from scratch with a dedicated focus on enhancing medical capabilities. Our model is trained on 20 trillion tokens and incorporates a range of effective training methods that strike a balance between general capabilities and medical expertise. As a result, Baichuan-M1 not only performs strongly across general domains such as mathematics and coding but also excels in specialized medical fields. We have open-sourced Baichuan-M1-14B, a mini version of our model, which can be accessed through the following links.
	\end{abstract}
	
	\def\github{\raisebox{-1.5pt}{\includegraphics[height=1.05em]{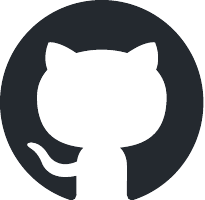}}}
	\def\huggingface{\raisebox{-1.5pt}{\includegraphics[height=1.05em]{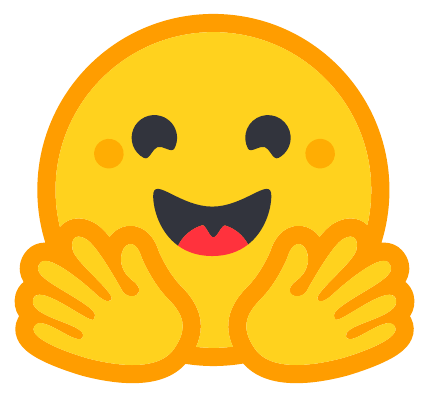}}}
	\newcommand{\ghlink}{https://github.com/baichuan-inc/Baichuan-M1-14B}
	\newcommand{\hfblink}{https://hf.co/baichuan-inc/Baichuan-M1-14B-Base}
	\newcommand{\hfilink}{https://hf.co/baichuan-inc/Baichuan-M1-14B-Instruct}
	\begin{center}
		\small
		\begin{tabular}{rl}
			\github & \href{\ghlink}{\textcolor{blue}{\ghlink}}\\
			\huggingface & \href{\hfblink}{\textcolor{blue}{\hfblink}}\\
			\huggingface & \href{\hfilink}{\textcolor{blue}{\hfilink}}\\
		\end{tabular}
		\normalsize
	\end{center}
	
	\begin{figure}[ht]
		\centering
		\includegraphics[width=0.8\textwidth]{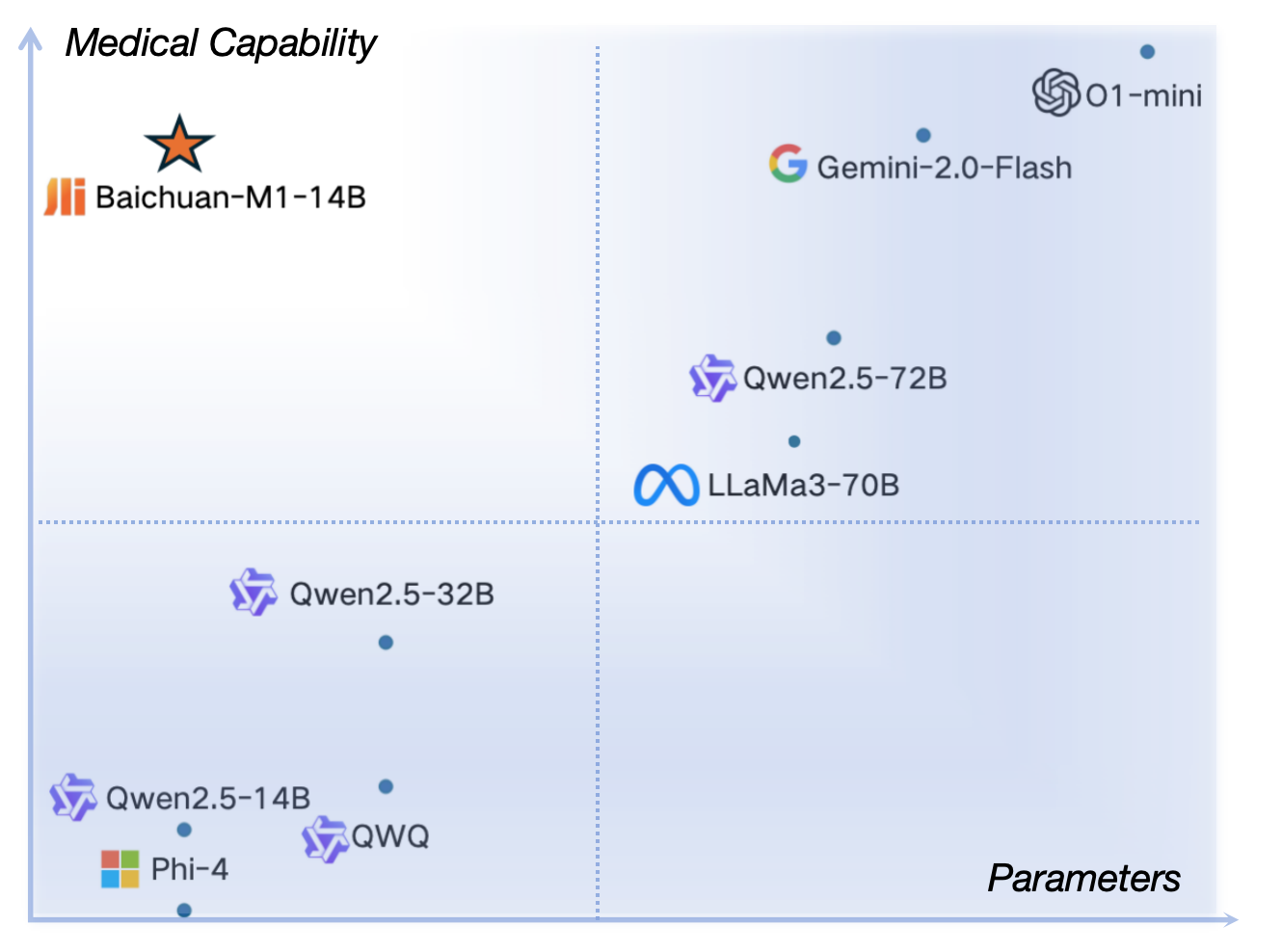}
		\caption{The medical capability of Baichuan-M1-14B compared with other models.}
		\label{fig:intro-mapping}
	\end{figure}
	
	\section{Introduction}\label{introduction}
	The rapid development of large language models (LLMs) has revolutionized the modern artificial intelligent by providing advanced capabilities in tasks such as natural language understanding, machine translation, text generation, and code generation. Sparked by the great success of ChatGPT \cite{chatgpt}, models like GPT-4 \cite{OpenAI2023GPT4TR} and Deepseek-R1 \cite{guo2025deepseek,deepseekai2024deepseekv3technicalreport} have demonstrated the vast potential of LLMs in handling general-purpose applications. These general-purpose models have opened up new possibilities across industries, from customer service to education \cite{adeshola2024opportunities,spurlock2024chatgpt}. However, despite their widespread success in broad domains, the application of LLMs in specialized fields, particularly vertical domains such as medical, remains underexplored and presents unique challenges.
	
	In the medical domain, the application of LLMs is both highly promising and profoundly challenging. Unlike more general applications, medical knowledge is complex, deeply specialized, and highly dynamic \cite{sanaei2023chatgpt,wang2024large}. Medical literature is vast, containing a wealth of information across a wide array of disciplines, including pharmacology, diagnostics, treatment protocols, and patient care. Furthermore, the language used in medical practice is often precise, technical, and sometimes ambiguous, requiring models to possess a high level of domain expertise to understand and generate contextually appropriate responses \cite{singhal2025toward,taylor2022galactica}. Traditional LLMs, while effective in general tasks, struggle to maintain this level of nuance and precision when applied to such specialized fields.
	
	A critical challenge in developing effective LLMs for the medical field is the availability of high-quality, domain-specific data. Medical data is often difficult to access due to privacy concerns, regulatory restrictions, and the complexity of medical terminologies. Moreover, the available data is often a mix of structured and unstructured information, encompassing clinical notes, research papers, medical textbooks, and more \cite{taylor2022galactica}. This makes it difficult to effectively pretrain models that can generalize across the full spectrum of medical knowledge. In response to these challenges, recent efforts have focused on fine-tuning general-purpose LLMs on medical datasets or applying transfer learning approaches \cite{10.1145/3686807,zhang2023huatuogpt,singhal2025toward,chen2024huatuogpt}. However, these methods often fail to capture the full depth and specificity of medical knowledge, resulting in models that may perform well in certain tasks but lack the expertise needed for more complex medical queries.
	
	In this paper, we introduce Baichuan-M1, a new series of large language models specifically trained for medical applications. Unlike traditional approaches that rely on continuing pretraining or post-training on general models, Baichuan-M1 is built from the ground up with a dedicated focus on medical expertise. The model is trained on 20 trillion tokens, including general data such as code and books, and medical related data such as clinical data, and patient-related information, using techniques that balance general language capabilities with specialized medical knowledge. This approach ensures that Baichuan-M1 excels not only in general areas like mathematics and coding, but also in specialized medical tasks such as diagnostic support, medical research, and treatment recommendations. Additionally, we introduce an enhanced Transformer architecture taking a balance between efficiency and effectiveness. Moreover, we incorporate a sophisticated training process that progressively refines the model's medical capabilities, ensuring continuous improvement in its performance on medical but also the general domain.
	
	The Baichuan-M1 series represents a significant step forward in the development of large language models for healthcare. By focusing on a vertical domain like medicine, we aim to push the boundaries of what LLMs can achieve in fields where precision and reliability are critical. This paper outlines the key features of Baichuan-M1, including its architecture, training methodology, and performance across both general and specialized tasks. We also present comparative results that demonstrate the model's ability to tackle medical queries with higher accuracy and relevance than general-purpose LLMs, marking a major leap toward the next generation of medical AI.

	\section{Data}
	
	Our pre-training process consists of several key components. First, we carefully design filtering and scoring mechanisms, leveraging the Baichuan series models to select over 20T tokens of high-quality training data. Second, we classify the data into different domains and apply strategic data mixing methods. We dedicate significant effort to collecting detailed medical data and design a three-stage pre-training process that gradually increases the proportion and complexity of medical data while extending the context window to enhance long-context understanding. Third, we meticulously develop a synthetic data strategy for high-quality medical data, generating over 100 billion high-quality medical reasoning tokens. Below, we provide a detailed explanation of our data preparation and training methodology.
	\subsection{General Data}\label{sec:general-data}
	Powerful medical capabilities are inseparable from strong general capabilities. We develop a multilingual general dataset with a total size of 20T tokens. The specific proportions of different languages are detailed in Table \ref{tab:dataset_composition}, covering the top 30 mainstream languages worldwide. The improvements of general data stem from several key aspects:
	
	
	\begin{table}[h]
		\centering
		\begin{tabular}{lccccc}
			\toprule
			\textbf{Language}    & English & Chinese & Multilingual & Code & Total \\
			\midrule
			\textbf{Token Count} & 12T     & 4T      & 2T           & 2T   & 20T  \\
			\bottomrule
		\end{tabular}
		\caption{Dataset Composition}
		\label{tab:dataset_composition}
	\end{table}

	\begin{enumerate}
		\item [(1)] \textbf{Global Deduplication and Upsampling Strategies.} Deduplication is a critical step in constructing high-quality training datasets for LLM pre-training, directly impacting both data quality and the effective training capacity of the model. We apply global deduplication across multilingual and multi-source data while recording the duplication count for each document, similar to Txt360~\citep{txt360}. 
		Documents without any duplicate matches may indicate lower quality or sparse distribution \citep{penedo2024finewebdatasetsdecantingweb}. To address this issue, we implement controlled upsampling based on the natural distribution of document duplication counts, ensuring a higher proportion of high-quality data.
		Figure~\ref{fig:data_deduplication} presents the performance curves of the model under the global deduplication and data upsampling strategies, where the deduplication curve represents the global deduplication strategy, and the "Deduplication + Upsampling" curve represents the strategy of first performing global deduplication and then upsampling based on document duplication counts. We trained a 3B model on approximately 1T tokens from scratch using an architecture similar to Baichuan M1. The results demonstrate that the Deduplication + Upsampling strategy achieves significantly better performance compared to global deduplication alone. Furthermore, compared to single-bucket deduplication, the Deduplication + Upsampling strategy provides better control over the total token count, allowing for a more balanced and effective training process.
		

		
	
	\begin{figure}[ht]
		\centering
		\includegraphics[width=0.7\textwidth]{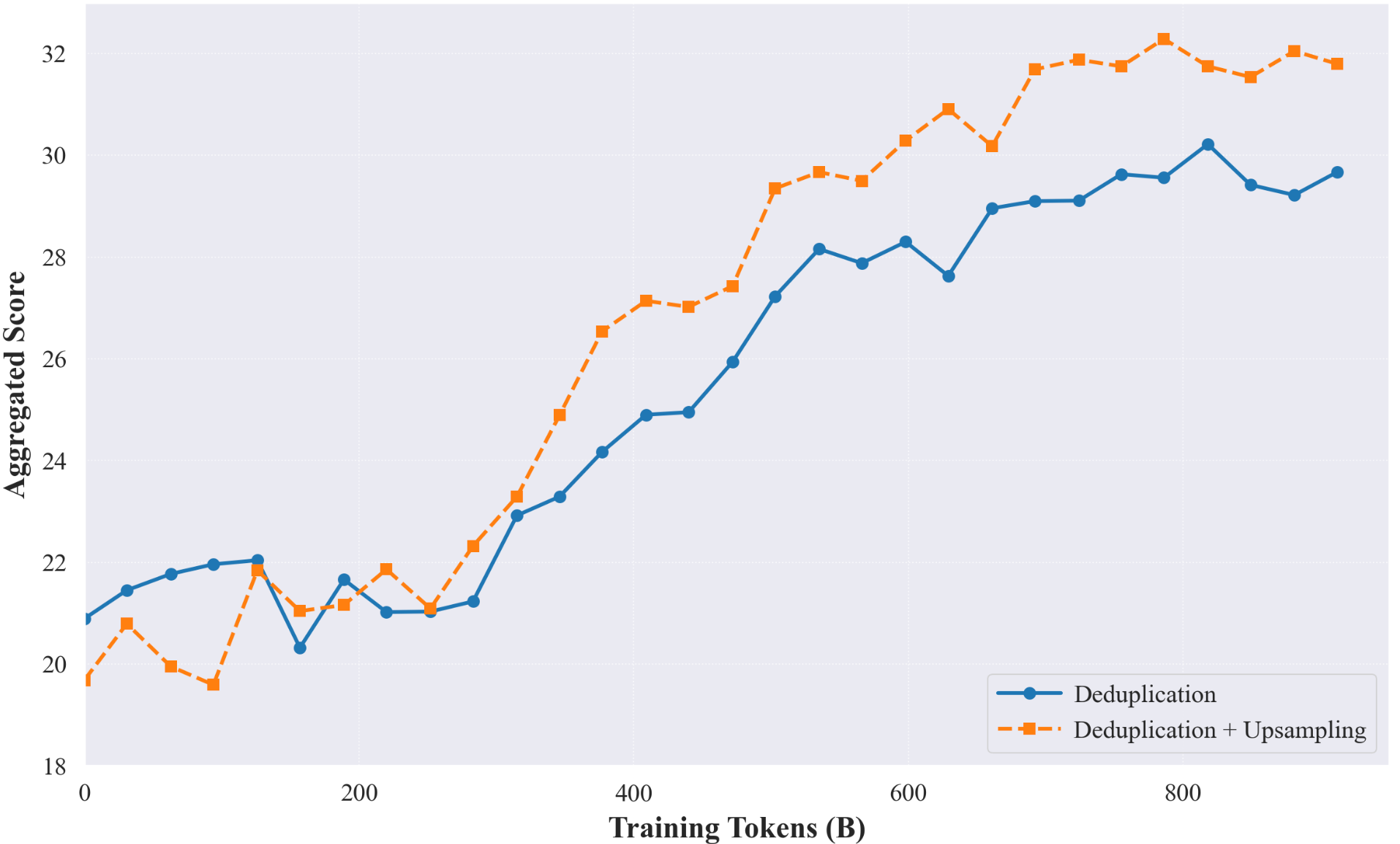}
		\caption{Global deduplication strategy, comparison with global deduplication + upsampling by count strategy. The Aggregated score represents the average performance across the following evaluation benchmarks: MMLU \citep{hendrycks2021measuring}, CMMLU \citep{li2024cmmlu}, GAOKAO \citep{zhang2024evaluatingperformancelargelanguage}, MBPP \citep{austin2021programsynthesislargelanguage}, JECQA \citep{zhong2019jecqalegaldomainquestionanswering}, USMLE \citep{jin2021disease}, and MCMLE \citep{jin2021disease}.}
		\label{fig:data_deduplication}
	\end{figure}
	
	\item [(2)]\textbf{Multidimensional Data Quality Assessment and Sampling Strategy.} 
	Enhancing data quality is essential for model training. We utilize a series of small models to perform multidimensional quality assessments on the entire dataset from both human and model perspectives, including metrics such as causal scoring, educational scoring, reasoning density, and knowledge density. 
	After obtaining the multidimensional quality scores for the data, we turn our attention to the next important challenge: determining the optimal sampling strategy.
	To determine the optimal sampling strategy, we conduct a detailed ablation study to explore the impact of different upsampling strategies on high-quality data. 
	Figure~\ref{fig:data_repetition} shows our comparative experiments on data quality and data quantity.
	Our findings indicate that appropriately upsampling high-quality data, compared to merely filtering out low-quality data, can significantly enhance overall model performance. Additionally, we observed that even aggressive upsampling strategies do not degrade model performance—for instance, upsampling the top 10\% high-quality data ten times yields comparable results to upsampling the top 33\% data three times.  
	To ensure the model is exposed to a diverse range of data, we constrain our upsampling strategy to a maximum of ten repetitions while removing low-quality data during pretraining.
	
	\textbf{For code data}, we use the Qwen2.5 Coder model\citep{qwen2.5_coder}, fine-tuned as a quality filter. 
	Under the condition of maintaining a consistent volume of code data, our comparison of different quality filtering strategies revealed that retaining the top 20\% of the highest quality code and applying 5x upsampling yields better results than simply filtering out low-quality data. Furthermore, this approach also outperforms using the top 50\% of the code data with 2x upsampling.
	Therefore, during the pre-training phase, we adopted the strategy of removing low-quality code data and upsampling the high-quality data.
	
	\textbf{For multilingual data}, we use a fine-tuned multilingual language model as the quality filter.
	We found that simply filtering out low-quality data and mixing in a large amount of multilingual data could not maintain 
	the model's original capabilities in Chinese and English.
	To ensure the model's capabilities in both Chinese and English and to utilize multilingual data more effectively, we conducted a series of experiments on multilingual data filtering strategies and mixing ratios.
	As shown in Figure~\ref{fig:data_multilingual}, we can observe that when selecting the highest quality multilingual data and mixing in 10\%, the multilingual data not only had no negative impact on the model's capabilities in Chinese and English but rather slightly enhanced these capabilities.
	This improvement may be related to the richer knowledge content of high-quality multilingual data. Additionally, the model gained multilingual capabilities, increasing the total token count by approximately 10\%.

	\begin{figure}[ht]
		\centering
		\includegraphics[width=0.7\textwidth]{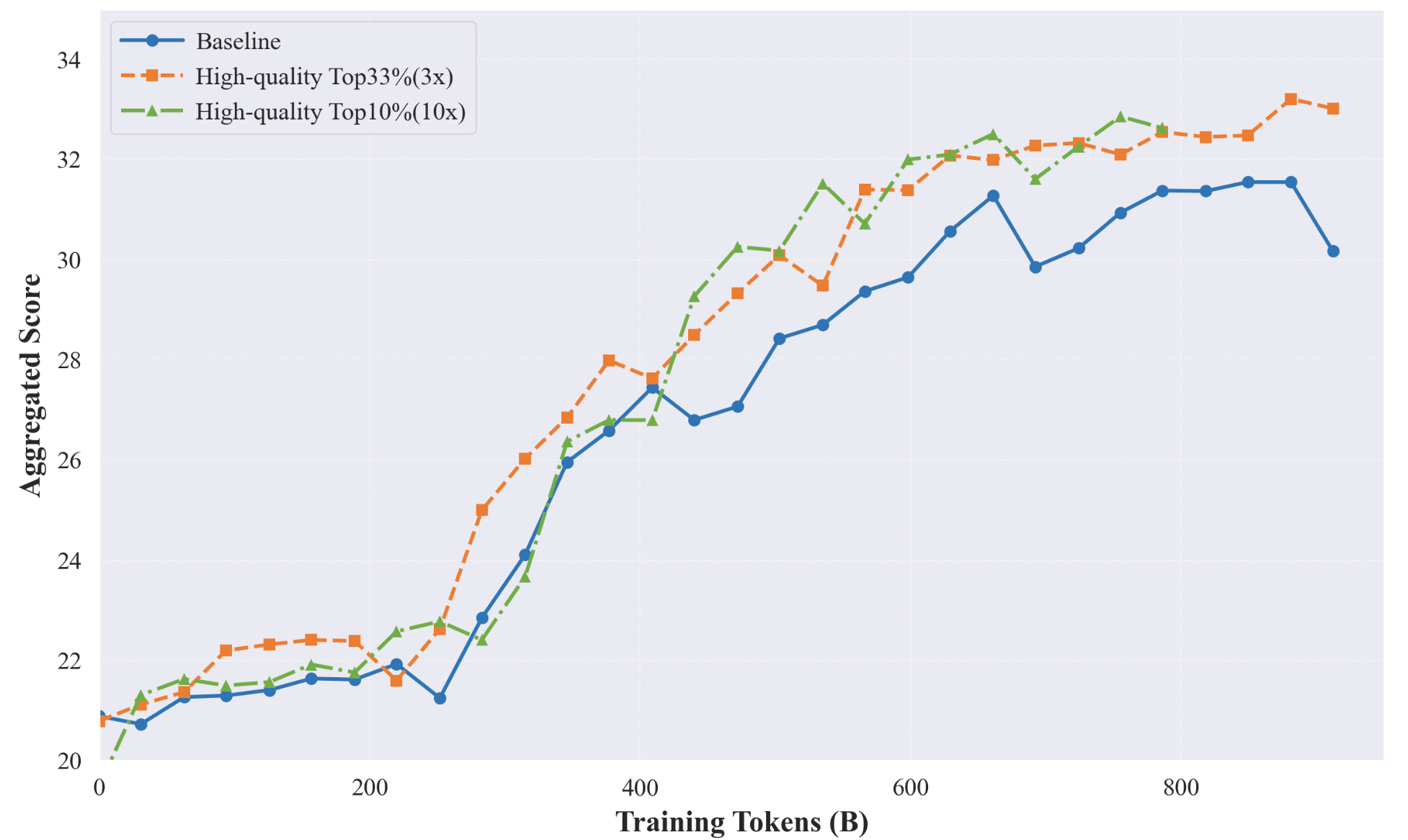}
		\caption{Comparison of data quality versus data volume. Baseline represents filtering out only low-quality data. High-quality Top 33\% (3x) refers to selecting the top 33\% highest-quality data and repeating it three times, while High-quality Top 10\% (10x) refers to selecting the top 10\% highest-quality data and repeating it ten times.}
		\label{fig:data_repetition}
	\end{figure}
	\begin{figure}[ht]
		\centering
		\includegraphics[width=0.7\textwidth]{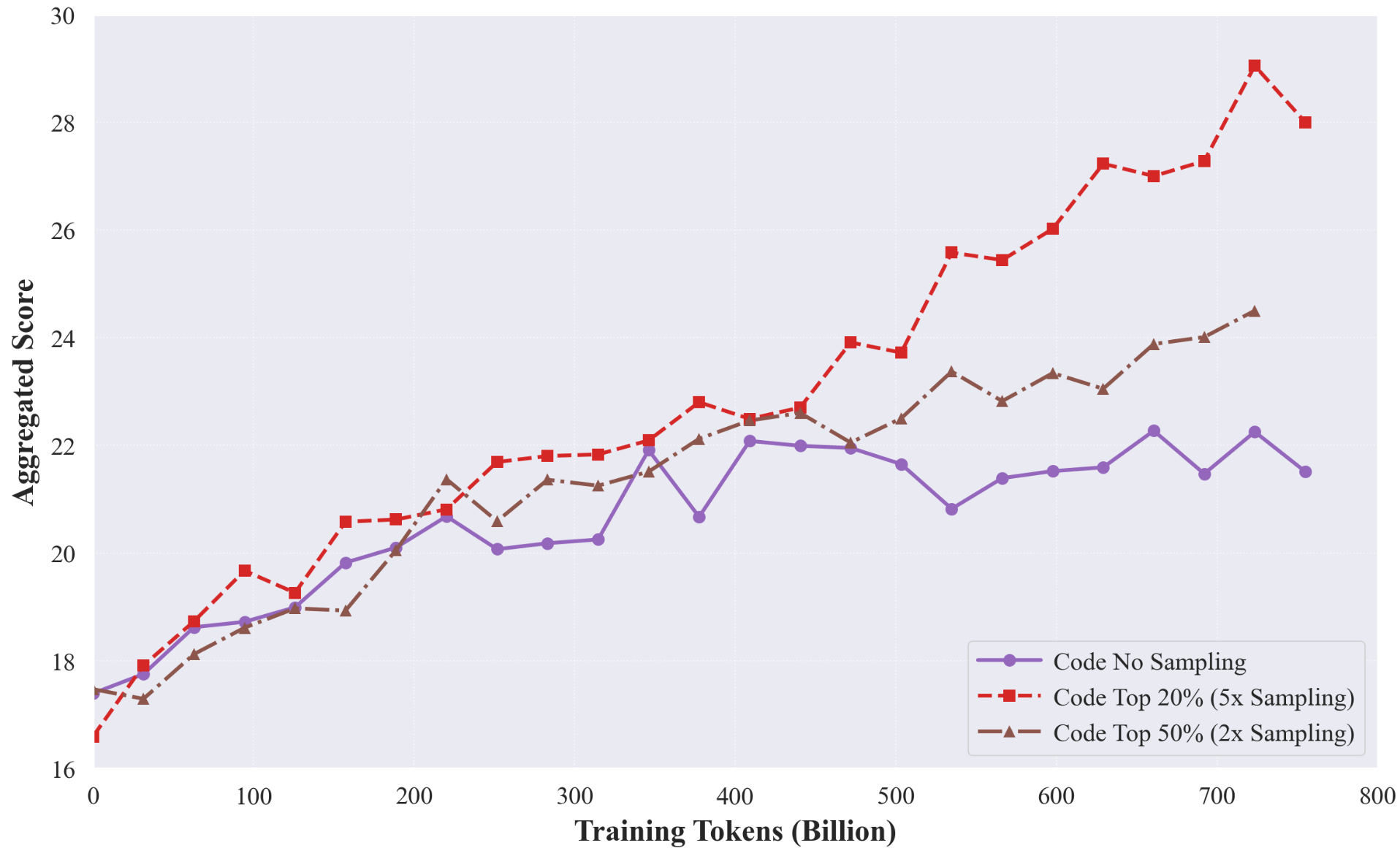}
		\caption{Comparison of different code data quality selection strategies and their impact on performance. Code No Sampling represents filtering out only low-quality data. Code Top 20\% (5x Sampling) refers to selecting the top 20\% highest-quality code data based on quality scores and repeating it five times. Code Top 50\% (2x Sampling) refers to selecting the top 50\% highest-quality code data and repeating it twice, while mixing in an equal proportion of general data. The Aggregated score represents the average performance across the following evaluation benchmarks: CEVAL \citep{huang2023ceval}, CMMLU, MMLU, GAOKAO, MBPP, and CMATH \citep{wei2023cmath}.}
		\label{fig:data_code quality}
	\end{figure}
	
	\item [(3)]\textbf{Data Classification and Ratio Optimization.} To explore better data ratios and optimize the distribution of pre-training data, we refer to the World Knowledge Classification System and use a series of small models to classify the data into 27 major categories. To ensure a balanced and information-rich dataset, we downsample overrepresented domains in the web-scale data, such as entertainment, e-commerce, news, and social media, while upsampling underrepresented domains like Science, Technology, Engineering, and Mathematics, which contain high-quality information. Additionally, we conduct large-scale data ratio experiments using a series of small models to fit the optimal data ratio strategy to maximize the model’s performance across various domains.
	\begin{figure}[ht]
		\centering
		\includegraphics[width=0.7\textwidth]{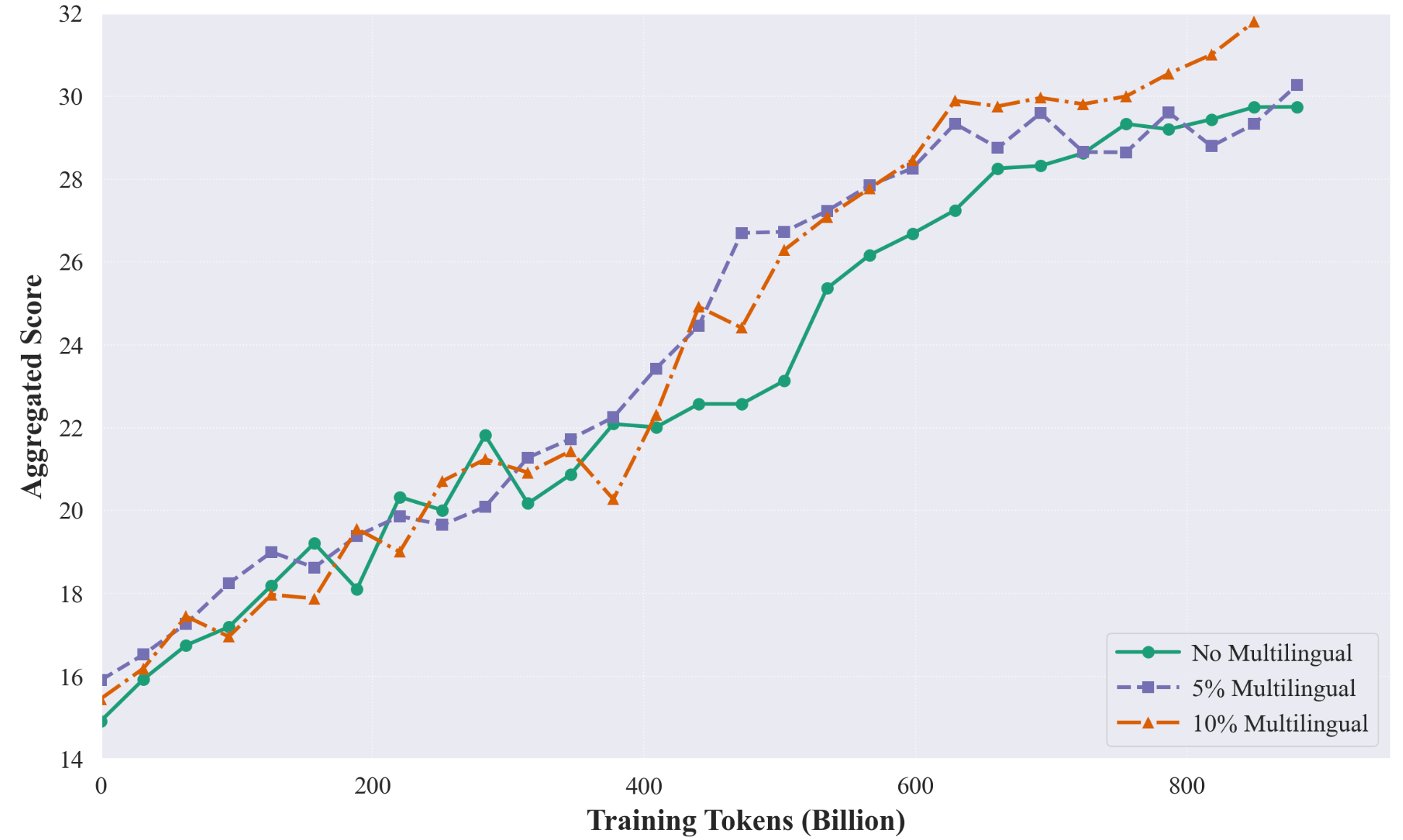}
		\caption{Comparison of different multilingual data filtering strategies and mixing ratios.}
		\label{fig:data_multilingual}
	\end{figure}
	\item [(4)]\textbf{Synthetic Data.} High-quality synthetic data, due to its step-by-step generation process and consistency with reasoning scenarios, significantly enhances the model's reasoning abilities and problem-solving skills \citep{abdin2024phi4technicalreport}. In domains such as mathematics, coding, and STEM, we leverage state-of-the-art models to synthesize large-scale data. To ensure high quality, we apply rigorous filtering using a general reward model trained on a language model. The synthetic data is utilized during the model annealing phase to further refine performance.
	\item [(5)]\textbf{Data Concatenation.} To preserve data integrity, we optimize the concatenation scheme to minimize unnecessary truncation of long sequences. Our experiments show that preventing truncation improves overall model performance and significantly enhances long-context understanding during training.
\end{enumerate}

\subsection{Medical data}\label{sec:medical-data}
To enhance the medical capabilities of our model, we dedicate significant efforts to collecting large-scale, high-quality, and diverse medical data for pre-training. 
We employ a sophisticated data classification and filtering process to improve data quality and ensure comprehensive coverage of medical domains. 
Furthermore, synthetic data has been demonstrated to be effective for model training \citep{abdin2024phi4technicalreport, chen2024diversitysyntheticdataimpact, maini2024rephrasingwebrecipecompute}, and thus, we generate extensive synthetic data to complement the medical data. 
Specifically, we utilize various advanced data synthesis techniques and design tailored synthesis strategies for different data sources, resulting in data of substantial scale, diverse formats, high educational value, and full coverage of medical scenarios. 
We conduct a three-stage training process, progressively increasing the concentration and complexity of medical knowledge. Using small-scale models for ablation studies, we determine the optimal data mixture strategy, enabling the model to transition from a medical student to a physician and ultimately to an expert.

\subsubsection{Data source}
The medical data are primarily obtained from two sources: classification and filtering of web corpora, and manually curated authoritative and informative sources.

\textbf{Web corpora.} We utilize the a series of small models to classify large-scale web data and further categorize medical data into specific medical content and department-based classifications. To ensure balanced data distribution, we downsample overrepresented categories, such as health and wellness advertisements. The classification of web-scale data facilitates the generation of a substantial volume of medical-related tokens.

\textbf{Expert-curated sources.} Due to the importance of authority and accuracy in medical content, coupled with the vast array of specialized topics of long-tailed knowledge within the medical domain, relying solely on web-based medical content is insufficient for training an expert-level medical LLM. 
To address this, we engage a team of medical experts to manually curate authoritative sources of medical knowledge, including medical academic papers, real-world medical cases, medical textbooks, biomedical knowledge graphs\footnote{To fully leverage structured knowledge from KGs for LLM pre-training, we transform it into natural language by anchoring each entity, incorporating its synonyms, semantic types, definitions, and related entities, and reorganizing the information using a Markdown template.
} (KGs), clinical guidelines, medical encyclopedias, and online medical customer QA data.
With extensive efforts from medical experts, complied the most comprehensive medical database to date, encompassing over 1T tokens from over 200 authoritative medical knowledge sources. This database almost covers the entire spectrum of medical knowledge, from macro to micro levels, and spans the complete system of medical scenarios, from cutting-edge research to clinical practice.

\textbf{Data quality filter.} Data quality filtering has been shown to be a critical component in the data collection pipeline \citep{qwen2025qwen25technicalreport, deepseekai2024deepseekv3technicalreport}. To enhance the quality of medical data, we specifically design a medical quality score and a medical value score. Using the a series of models, we score all medical data and perform sampling based on these scores, significantly improving the overall quality of the medical dataset.

\subsubsection{Synthetic Data}
\begin{figure}[ht]
	\centering
	\includegraphics[width=0.9\textwidth]{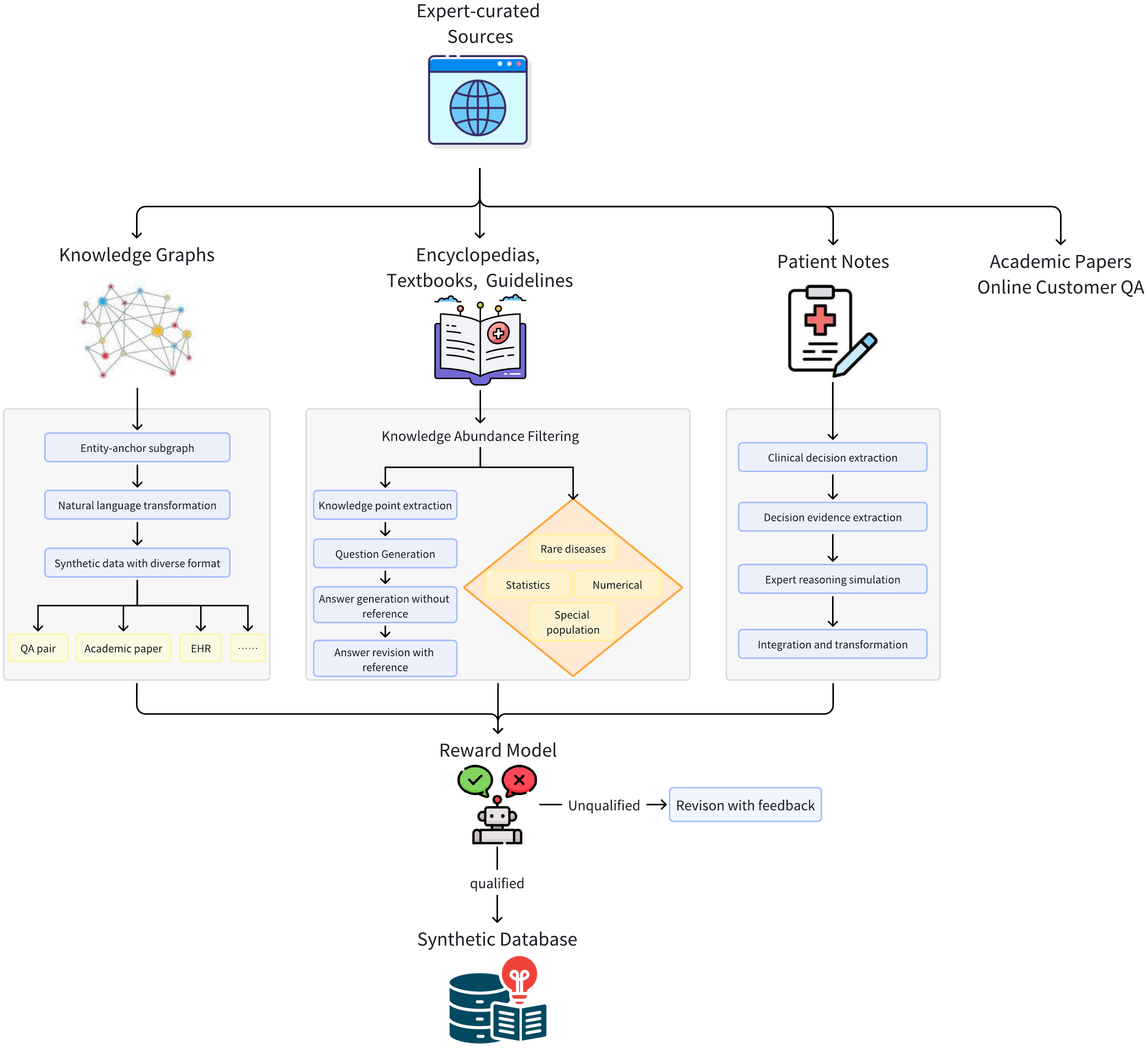}
	\caption{The data synthesis pipeline for various sources.}
	\label{fig:synthetic_data}
\end{figure}

The use of synthetic data in LLM training has gained increasing popularity and has even become a substantial component of training datasets \citep{abdin2024phi4technicalreport}. 
To enhance the quality of synthetic medical data, we manually design distinct yet sophisticated synthesis pipelines tailored to different data sources. 
Specifically, we utilize expert-curated sources after rigorous filtering as seed data to ensure both the accuracy of synthesis and a comprehensive coverage.
Furthermore, we emphasize the importance of long Chain-of-Thought (CoT, \citealp{Wei2022ChainOT}) and actively encourage its generation for all synthesized data.
We leverage state-of-the-art LLMs to generate data.
The detailed examples of synthesis pipelines for various sources are provided below and summarized in Figure \ref{fig:synthetic_data}.

\textbf{Encyclopedias, textbooks, and guidelines.} These sources contain extensive medical knowledge presented in natural language.
Following \citet{abdin2024phi4technicalreport}, we construct QA pairs to enhance the medical knowledge of our model.
Concretely, we first split long documents (potentially millions of tokens) into shorter chunks (thousands of tokens) and perform an additional round of filtering focused on "knowledge abundance".
This step leverages a language model to eliminate chunks that do not demonstrate specific medical knowledge points.
Subsequently, we conduct data synthesis using the following refined pipeline:
\begin{enumerate}
	\item \textbf{Knowledge point extraction}: The model extracts medical knowledge points (scientific facts) from the documents.
	\item \textbf{Question generation}: For each knowledge point, the model generates an exam question, which can be either a multiple-choice question or a short-answer question.
	\item \textbf{Answer generation without reference}: For each generated question, the model provides an answer without referring to the original document, with an emphasis on generating long Chain-of-Thought (CoT) reasoning. 
	This step preserves the LLM's output patterns and full CoT structure, facilitating the student model's learning.
	\item \textbf{Answer revision with reference}: Since the initially generated answers may be incorrect, we conduct a revision step where the model revises its answer based on the original document.
\end{enumerate}
By combining these steps, we obtain accurate QA pairs covering a wide range of medical topics, accompanied by comprehensive CoT reasoning that aligns with the model's learning process.
We additionally focus on long-tailed medical knowledge including biomedical statistics, numerical problems, rare diseases, and special populations.

\textbf{Real-world patient cases.} Real-world patient notes are invaluable resources for models to learn clinical practices and develop expertise in the medical domain. 
However, patient notes typically omit the detailed reasoning processes of physicians, recording only final decisions such as diagnoses, treatments, and examinations. 
To address this limitation, we propose the following pipeline to reconstruct an expert-level reasoning process based on real-world cases:
\begin{enumerate}
	\item \textbf{Clinical decision extraction}: The model identifies critical clinical decisions made throughout the patient's journey, from preliminary diagnosis to prognosis prediction.
	\item \textbf{Decision evidence extraction}: For each clinical decision, the model extracts all relevant evidence, including both positive and negative manifestations that support the decision.
	\item \textbf{Expert reasoning simulation}: Using the extracted evidence, the model simulates the reasoning process of a medical expert, with particular emphasis on evaluating alternatives such as differential diagnoses and treatment options.
	\item \textbf{Integration and transformation}: The model integrates the components above, transforms the original note (particularly the evidence) into a suitable format, and weaves together a comprehensive reasoning process that mirrors the full thought process of medical experts.
\end{enumerate}

\textbf{Knowledge graphs.} For each entity in KGs, we have transformed related knowledge into natural language using a Markdown template.
Based on this, we generate data in multiple formats following \citet{maini2024rephrasingwebrecipecompute} to expose the model to diverse knowledge representations. 
Notably, during QA pair generation, we fully leverage the structure of KGs and propose the synthesis strategy of "entity as the answer".
Similar to \citet{chen2024reversethinkingmakesllms}, we encourage the model to generate backward reasoning questions by anchoring an entity as the answer rather than including it in the question. 
For example, for a disease entity, we encourage questions such as "Which of the following diseases could cause the symptom of ...?" instead of "What are the typical symptoms of ... disease?" 
Additionally, the abundant relationships in KGs often cluster similar entities that require careful differentiation. 
Therefore, we utilize these similar entities as options for multiple-choice questions and generate questions focusing on subtle differences among them, such as "Which of the following vitamins is provided only by animal sources?"

\textbf{Academic Papers.} Academic papers encapsulate cutting-edge advancements in the medical domain, characterized by rigorous reasoning and a formal tone. To enhance the model’s scientific reasoning capabilities, we extract key evidence and conclusions from these papers and leverage the state-of-art model to generate detailed analytical bridges between them. This approach enables the student model to learn and internalize the scientific deduction process, improving its ability to engage in structured reasoning.

\textbf{Online customer QA.} Online customer QA data also plays a critical role in model training.
Customer questions often contain nonstandard terminology and corner cases that may be overlooked in authoritative sources.
However, answers from online forum, even those written by physicians, also suffer from the usage of informal language and excessive brevity.
Therefore, we first ask the model to answer the question on itself, and then revise its answer by referring to the original answer to achieve a balance between accuracy and comprehensiveness.
\\\\
The synthetic data also undergo rigorous quality check with the medical value reward model.
Qualified data is directly used as training data, while unqualified data is revised based on feedback from the reward model until it meets satisfactory quality standards.

\section{Model Architecture}
Overall, our structure is similar to Llama and other popular models
\citep{touvron2023llama,yang2023baichuan,bai2023qwen}, including the use of a pre-norm based on rmsnorm \citep{vaswani2017attention,xu2019understanding,zhang2019root}, FFN layer using SwishGlu \citep{shazeer2020glu}, and rotary position embedding \citep{su2024roformer}. In addition, in order to reduce the inference cost, we alternately used global attention and sliding window attention like Gemma2 \citep{team2024gemma}. In fact, the proportion of sliding window attention can be appropriately increased \citep{yang2025rope}. We also increased the head dim from 128 to 256 for the global attention part of the model, as our early experiments find that a head dim of 256 is beneficial for the emergence of some benchmarks.

\begin{wrapfigure}{r}{0.6\textwidth}
	\centering
	\includegraphics[width=0.44\textwidth]{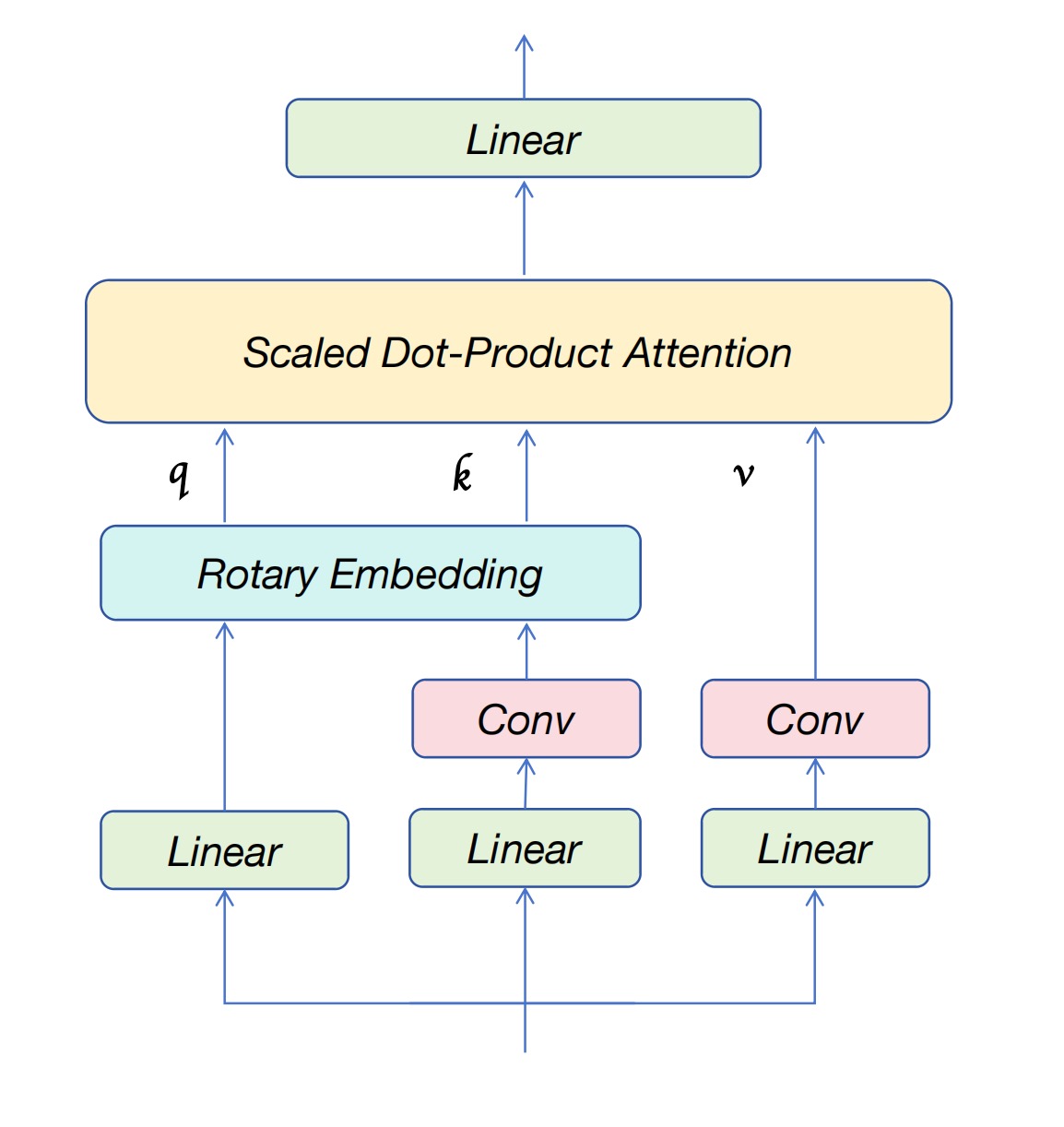}
	\caption{The attention mechanism used by Baichuan-M1-14B.}
	\label{fig:kv_shift_attention}
\end{wrapfigure}

In addition, we also used temporal short convolution operations on the key and value in attention as shown in Figure \ref{fig:kv_shift_attention}, which is beneficial for the formation of the model's in-context learning ability \citep{xu2024kv}. This structure has been found by neural network architecture searching \citep{so2021searching} in  language modeling, and it is also widely used in non transformer structures \citep{fu2022hungry,peng2023rwkv,beck2024xlstm,yang2024parallelizing}.

\begin{figure}[ht]
	\centering
	\includegraphics[width=0.75\textwidth]{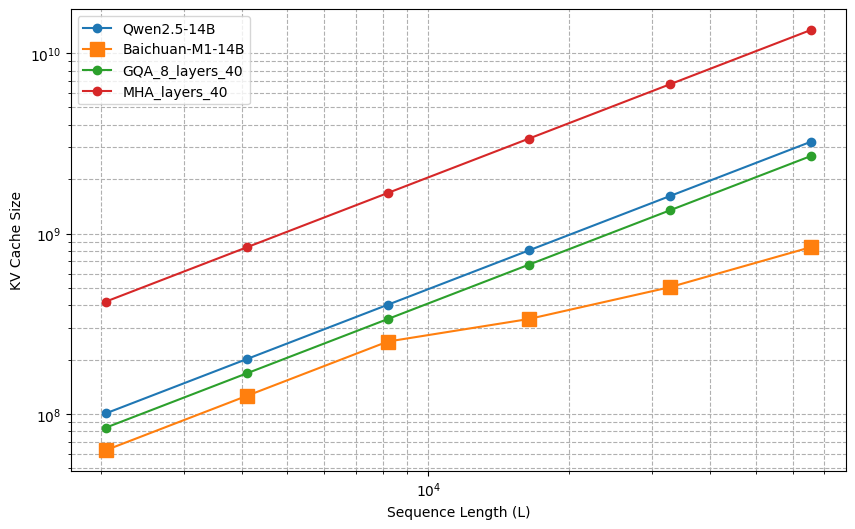}
	\caption{KV cache Comparison between Baichuan-Med-14B and other models. In the case of a short context, the KV cache of Baichuanmed-14B is approximately equal to GQA of 6 KV heads, and in the case of a long context, it is approximately equal to the GQA of 4 kv heads.}
	\label{fig:kvcache}
\end{figure}

We use base=1,000,000 for rotary embedding when pretraining rather than 10,000 like Llama2, because the context length of the model is 32k, which requires a large base \citep{xu2024base}. If we use a smaller base, both the benchmark for short context and the perplexity for long context will be fine, while the model will lose the ability for long-distance retrieval. 

And we conduct ablation experiments on the model with 1.5B parameters and trained with 200B tokens, the results are shown in Table \ref{tab:ablation}. The sliding window size is 2k, the total training length is 8k, and NIAH stands for needle in a haystack. And we evaluate our model on multiple commonsense reasoning benchmarks: PIQA \citep{bisk2020piqa}, HellaSwag (Hella) \citep{zellers2019hellaswag},
WinoGrande (Wino) \cite{sakaguchi2021winogrande}, ARC-easy (ARC-e) and ARC-challenge (ARC-c) \cite{clark2018think}, SIQA, BoolQ, Wikitext (Wiki), and LAMBADA (LMB). And we evaluate Wiki and LMA by ppl, Hella and Arc-E by norm accuracy, others by accuracy. 

In addition, to save KV cache and improve inference efficiency, we also alternate the use of sliding window attention. The global attention layer has 2 heads with a head dim of 256, while the sliding window attention layer has 8 heads with a head dim of 128. One important reason for adopting this interleaving structure is that large language models have a large amount of layer redundancy  
\citep{men2024shortgpt}, and the number of heads with long-term retrieval capabilities is relatively small \citep{wu2024retrieval}.

From Table \ref{tab:ablation}, it can be seen that mixing with sliding window attention does not significantly affect the performance of long context benchmark, but can improve the performance of short context benchmarks. This suggests that the hybrid model may have better performance, which is consistent with some previous research \citep{Waleffe2024AnES,yang2025gated}.

\begin{table}[h!]
	\small
	\centering 
	\setlength{\tabcolsep}{0.5mm}
	\begin{tabular}{l|cc|cccccccc|c}
		\toprule
		\textbf{Model} & \textbf{Wiki} $\downarrow$ & \textbf{LMB} $\downarrow$ & \textbf{PIQA} $\uparrow$ & \textbf{Hella} $\uparrow$ & \textbf{Wino} $\uparrow$ & \textbf{ARC-e} $\uparrow$ & \textbf{ARC-c} $\uparrow$ & \textbf{SIQA} $\uparrow$ & \textbf{BoolQ} $\uparrow$ & \textbf{Avg}$\uparrow$ & \textbf{Niah}$\uparrow$ \\
		\hline
		Baichuan & \textbf{15.39} & 14.93  & 79.01 & 61.70 & 63.47 & 77.92 & 43.46 & 45.81 & 66.40 & 62.54 & \textbf{93.6} \\
		H.D.=128 & 15.72 & 15.67  & 77.41 & 60.63 & 62.17 & 77.04 & 40.59 & 43.38 & 63.20 & 60.63 & 92.3 \\
		75\% swa & 15.43 & \textbf{14.87}  & 79.12 & 60.16 & 64.59 & 78.02 & 43.24 & 46.21 & 66.64 & \textbf{62.57} & 89.4 \\
		w/o swa & 17.18 & 16.28  & 76.72 & 60.77 & 64.00 & 72.86 & 42.62 & 45.97 & 63.56 & 60.93 & 93.3 \\
		w/o conv & 17.97 & 16.96 & 77.04 & 58.71 & 60.75 & 76.15 & 39.32 & 42.80 & 63.34 & 59.73 & 88.4 \\
		base=1e4 & 15.67 & 15.03  & 78.61 & 61.60 & 61.15 & 79.24 & 42.92 & 45.36 & 66.29 & 62.02 & 91.2 \\

		\bottomrule
	\end{tabular}
	\caption{Ablation experiments we conducted on the model with 1.5B parameters and trained with 200B tokens. H.D. is short for head dim.}
	\label{tab:ablation}
\end{table}

\section{Training Progress}
\subsection{Tokenizer} The medical domain is characterized by a vast array of specialized terminology. Previous studies have demonstrated that training large language models (LLMs) for the medical domain requires a domain-specific vocabulary to optimize performance \cite{taylor2022galactica,shin2020biomegatron}. In the case of Baichuan-M1, we build the vocabulary from two perspectives: first, we train a general-purpose tokenizer using a broad general-purpose dataset such as web pages and books. Next, we develop a specialized tokenizer focused on medical terminology. Finally, we combine these two tokenizers to achieve a balance between general tokenization efficiency and the granularity needed for medical content. Our final vocabulary size is 133,120. The token efficiency of Baichuan-M1, compared to other models, is illustrated in Figure \ref{fig:tokenizer}. More details of our tokenizer is illustrated in Appendix.

\begin{figure}[ht]
	\centering
	\includegraphics[width=0.65\textwidth]{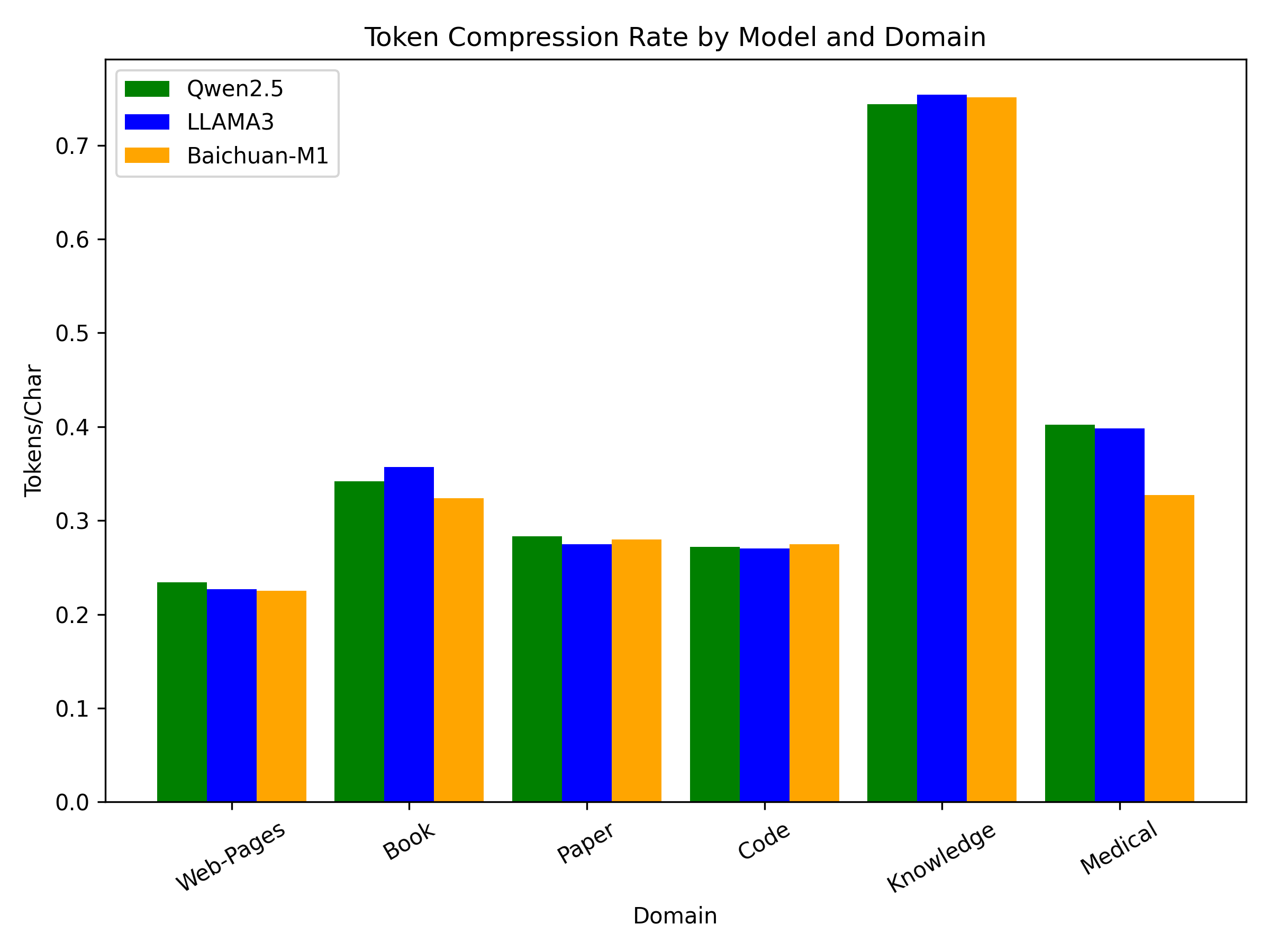}
	\caption{The tokenization efficiency for different models, the less the better.}
	\label{fig:tokenizer}
\end{figure}

\subsection{Training details}

The pre-training consists of three distinct stages, following a curriculum learning approach. In the first stage, we select data based on multiple factors that are relatively \textit{easy}, as the model is randomly initialized, and we believe the training process should be progressive. We choose data samples that are easy both in terms of quality and perplexity. During this stage, we do not significantly increase the proportion of medical data. We use a previous version of our Baichuan model, which has the same data distribution with Baichuan-M1, to obtain these two metrics. 
In the second stage, we gradually increase the proportion of both \textit{hard} samples and medical data, as we believe that medical capabilities build upon general capabilities. The third stage is an annealing phase, where we introduce the most complex and application-specific data to prepare the model for downstream alignment.

For Baichaun-M1-14B, we employ the AdamW optimizer \cite{loshchilov2017decoupled} during training. The parameters $\beta_1$ and $\beta_2$ are set to 0.9 and 0.95, respectively. We apply a weight decay of 0.1 and clip the gradient norm to 1.0. The model undergoes 2,000 linear scaling steps for warm-up, gradually reaching the maximum learning rate. The learning rate schedule follows the warm-up-stable-decay strategy \cite{hu2024minicpm}, with a peak learning rate of \textit{4e-4}. The batch size is set to 16M. After training for 12 trillion tokens, we enter the second stage, where the proportion of higher-quality data is increased, along with a greater emphasis on serious medical data. During this phase, the learning rate and batch size remain unchanged, and training continues for an additional 6 trillion tokens. Following a total of 18 trillion tokens, we begin annealing, employing a cosine annealing strategy to gradually reduce the learning rate to a minimal value. After 2 trillion tokens of annealing, the learning rate reaches \textit{2e-5}, at which point we transition to SFT training. The context length was set to 8K in the first and second stage, and increased to 32K in the third (annealing) stage. The base value of RoPE was first set to 1e5 and then increased to 1e6.

\begin{figure}[ht]
	\centering
	\includegraphics[width=0.75\textwidth]{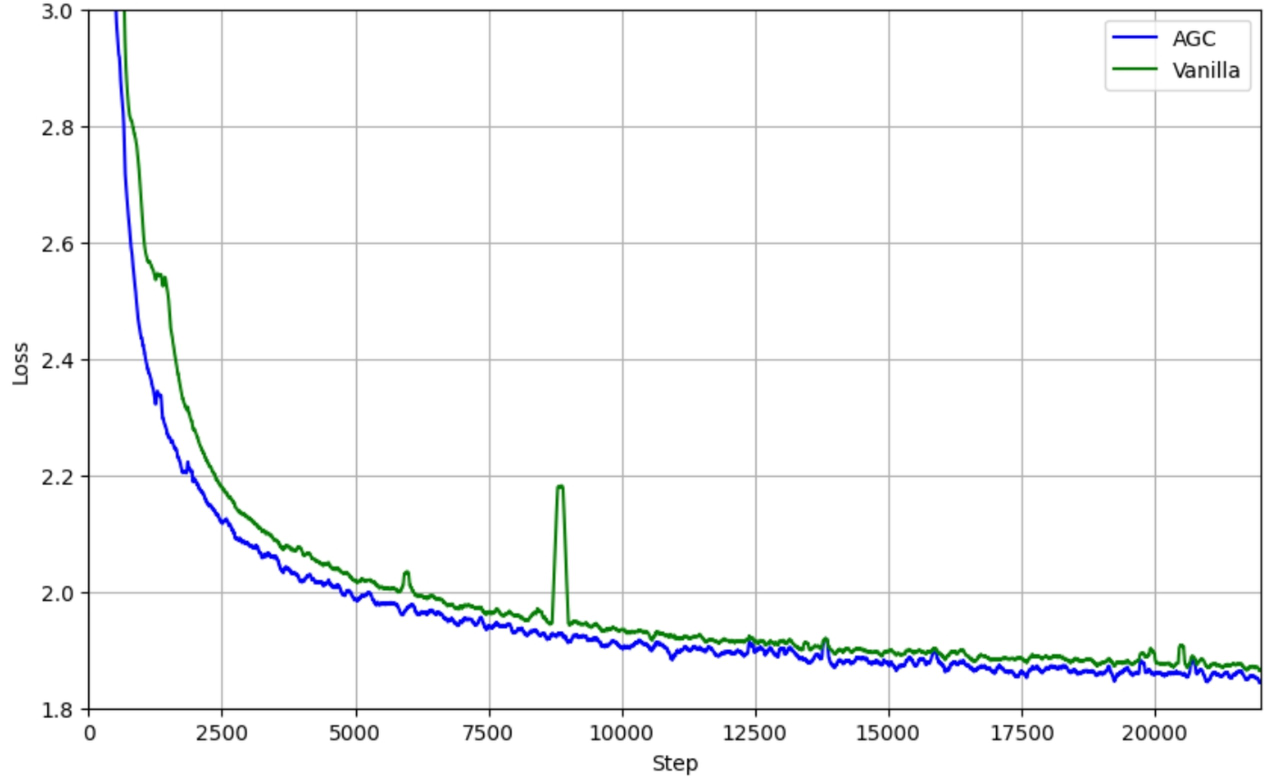}
	\caption{The Adaptive Gradient Clipping (AGC) strategy improves the stability of the initial loss during training.}
	\label{fig:agc}
\end{figure}

In the early stage of training, in order to stabilize the training, we adopted an adaptive gradient truncation strategy. The motivation is that during training, sometimes the large gradient is due to the current parameter reaching a steep point in the parameter space, and sometimes it is caused by special data, and we hope to eliminate the influence of the latter. Pseudo code as shown in Algorithm \ref{alg:gradient_update}. We conducted experiments on the 3b parameter model and demonstrated the effectiveness of this method in improving the stability of the model during early training, as shown in Figure \ref{fig:agc}.

\begin{algorithm}[H]
	\caption{Adaptive Gradient Clipping (AGC)}\label{alg:gradient_update}
	\SetKwInOut{Input}{Input}
	\SetKwInOut{Output}{Output}
	
	\SetKwData{Stack}{S}
	\SetKwData{SkipCounter}{skip\_counter}
	\SetKwData{MaxSkip}{MAX\_SKIP}
	\SetKwData{CurrentNorm}{current\_norm}
	\SetKwData{AvgNorm}{avg\_norm}
	
	\SetKwFunction{ComputeNorm}{compute\_norm}
	\SetKwFunction{UpdateParameters}{update\_parameters}
	
	Initialize an empty stack \Stack \tcp*{Store the last 100 gradient norm }
	\SkipCounter $\gets 0$ \tcp*{Initialize skip counter}
	\MaxSkip $\gets 1$ \tcp*{Set maximum consecutive skips}
	
	\For{each training step}{
		\CurrentNorm $\gets$ \ComputeNorm{} \tcp*{Compute the norm of the current gradient}
		
		\If{size of \Stack $\geq 100$}{
			\AvgNorm $\gets \frac{\sum_{i=1}^{100} \Stack_i}{100}$ \tcp*{Compute the average norm of the last 100 gradients}
			
			\If{\CurrentNorm $> 1.2 \times \AvgNorm + 0.1$}{
				\If{\SkipCounter $<$ \MaxSkip}{
					\SkipCounter $\gets$ \SkipCounter $+ 1$
					\tcp*{Skip step}
					\textbf{continue}
				}
				\Else{
					\SkipCounter $\gets 0$ \tcp*{Force parameter update}
				}
			}
		}
		
		\UpdateParameters{} \tcp*{Update model parameters}
		Push \CurrentNorm onto \Stack \tcp*{Store the current norm}
		
		\If{size of \Stack $> 100$}{
			Remove the oldest element from \Stack \tcp*{Maintain stack size}
		}
		
		\SkipCounter $\gets 0$ \tcp*{Reset skip counter}
	}
\end{algorithm}

\section{Alignments}

\subsection{Supervised Fine-Tuning}

We meticulously curated an instruction fine-tuning dataset comprising both a general instruction dataset and a medical instruction dataset.

\subsubsection{General SFT data}
For the general instruction dataset, we iteratively refined our in-house data and specifically constructed task-focused data, including mathematics and coding competitions, to emphasize reasoning skills.

\subsubsection{Medical SFT data}
Regarding medical alignment data, we subdivide it into five major categories to ensure comprehensive coverage of diverse clinical scenarios: 
\textbf{(1) Medical Knowledge, (2) Medical Language Understanding, (3) Medical Reasoning, (4) Medical Long Context,} and \textbf{(5) Medical Safety.} 
Below, we outline each category in detail.

\paragraph{Medical Knowledge Data}
We leverage online and open-source medical datasets to ensure a diverse range of inputs, which broadens the model's factual base. 
A variety of publicly available databases, research publications, and anonymized clinical notes contribute to robust coverage of diseases, treatments, and medical terminologies. 
This diversity in raw data is vital for training high-performing medical LLMs, as it reflects real-world medical complexity.

\paragraph{Medical Language Understanding Data}
To accurately interpret domain-specific texts and dialogues, we develop specialized tasks:
\begin{itemize}
	\item \textbf{Entity Extraction and Classification:} From patient-physician dialogues to structured medical records, the model is tasked with identifying key entities (e.g., symptoms, medications) and classifying them into standard categories.
	\item \textbf{Terminology Standardization:} We curate datasets of ``non-standard term''--``standard term'' pairs. This helps align the model output with accepted medical taxonomies, ensuring consistency and clarity.
	\item \textbf{Intent and Context Analysis:} Through classification and discourse tasks, the model learns to handle different forms of clinical text, such as Q\&A pairs, discharge summaries, and research articles.
\end{itemize}

\paragraph{Medical Reasoning Data}
Medical reasoning is essential for clinical decision-making. We implement a multi-stage pipeline to guide logical steps and quality assurance:
\begin{itemize}
	\item \textbf{Case Summaries and Differential Diagnosis:} The model synthesizes patient records and symptom data, prioritizing possible diagnoses and highlighting key clinical factors.
	\item \textbf{Treatment Recommendations and Prognosis:} Based on relevant guidelines, the model proposes therapeutic strategies, discussing their pros and cons, and offering prognostic insights.
	\item \textbf{Iterative Review Process:} The model undergoes multiple reasoning passes, refining its conclusions based on clinical best practices and additional feedback loops.
\end{itemize}

\paragraph{Medical Long Context Data}
Clinical data often includes lengthy and complex references (e.g., guidelines, manuals, long-form case reports). 
We address this by:
\begin{itemize}
	\item \textbf{Diverse Data Sources:} Medical manuals, guidelines, research papers, and extended patient records form the core corpus.
	\item \textbf{Task-Oriented Prompt Construction:} We create tasks such as question answering, summarization, rewriting, and evidence extraction. 
	\item \textbf{Context Extension:} Randomly concatenating additional documents before and after the main text expands the context window, training the model to manage long-sequence reasoning more effectively.
\end{itemize}

\paragraph{Medical Safety Data}
Ensuring medical safety involves both data diversity and robust quality control over prompts and model outputs.

\begin{itemize}
	\item \textbf{Define Coverage Labels:} We establish a comprehensive safety taxonomy, mapping potential high-risk topics or attack vectors to specific labels.
	\item \textbf{Set Safety Principles:} Each label is assigned clear guidelines on acceptable behavior and ethical considerations.
	\item \textbf{Refine Attack Strategies:} Based on accumulated knowledge of adversarial queries, we adjust and expand the database of potential attack types.
	\item \textbf{Illustrative Cases:} For each attack vector, we create 2--3 concrete examples to demonstrate how prompts could compromise patient privacy or safety.
	\item \textbf{Prompt Generation:} Using prompt engineering (PE) and few-shot learning, we produce the required number of prompts for each attack type within each category.
	\item \textbf{Deduplication:} Employing prefix/suffix alterations and semantic similarity checks, we remove duplicates to finalize the safety prompt set.
\end{itemize}

\paragraph{Answer Quality}
\begin{itemize}
	\item \textbf{Data Augmentation:} We rely on real-world patient cases, such as MIMIC \citep{johnson2016mimic, johnson2023mimic} and PMC-Patient \citep{zhao2023large}, as primary data sources to generate questions with sufficient difficulty, real-world complexity, and coverage of the full spectrum of clinical scenarios.
	\item \textbf{Candidate Answer Generation:} Combining each label’s safety principles with advanced LLMs, we produce multiple candidate answers for each prompt.
	\item \textbf{Human Expert Validation:} Domain experts review and refine the candidates to ensure they meet safety and accuracy standards, preserving critical medical context.
\end{itemize}

Each domain employed unique data creation strategies to fulfill its specialized requirements. We utilized both in-house and open-source models to generate the data. All original data sources, such as proprietary medical texts and patient records, underwent rigorous privacy protection measures. For reasoning-focused data, we adopted the format “system prompt + problem + response,” yielding entries in the form (\texttt{system prompt, problem, response}).

Finally, we fine-tuned Baichuan-M1-14B-Base over five rounds. During training, we used a cosine decay learning rate schedule starting at \texttt{2e-5} and gradually decreasing. When packing multiple samples into a single training sequence, we employed a “sample masking” strategy to prevent any cross-contamination between samples and ensure their independence.

\subsection{Reinforcement Learning}
\subsubsection{Reward Model}
We use both rule-based reward model (RM) and model-based reward model in our reinforcement learning process. \\
\textbf{Rule-based RM}\quad For questions with verifiable answers, we construct a rule-based system to obtain the quantifiable feedback. Rule-based RM can indeed cover a massive range of domains and applications. For example, ground-truth of medical diagnosis can be extracted from medical textbooks, case histories etc. For self-contained coding problems such as those on Leetcode, we can leverage compiler feedback of test cases to verify the correctness. The main advantage of rule-based RM is its high reliability.\\
\textbf{Model-based RM}\quad For questions with uncertain answers, we rely on a model-based RM (trained from a Baichuan-M1 checkpoint) to provide the feedback. When training the model-based RM, we meticulously construct a preference dataset to impose expert priors in different domains, to reduce the risk of reward hacking. 
\subsubsection{Training Policies}
\begin{figure}[ht]
	\centering
	\includegraphics[width=0.95\textwidth]{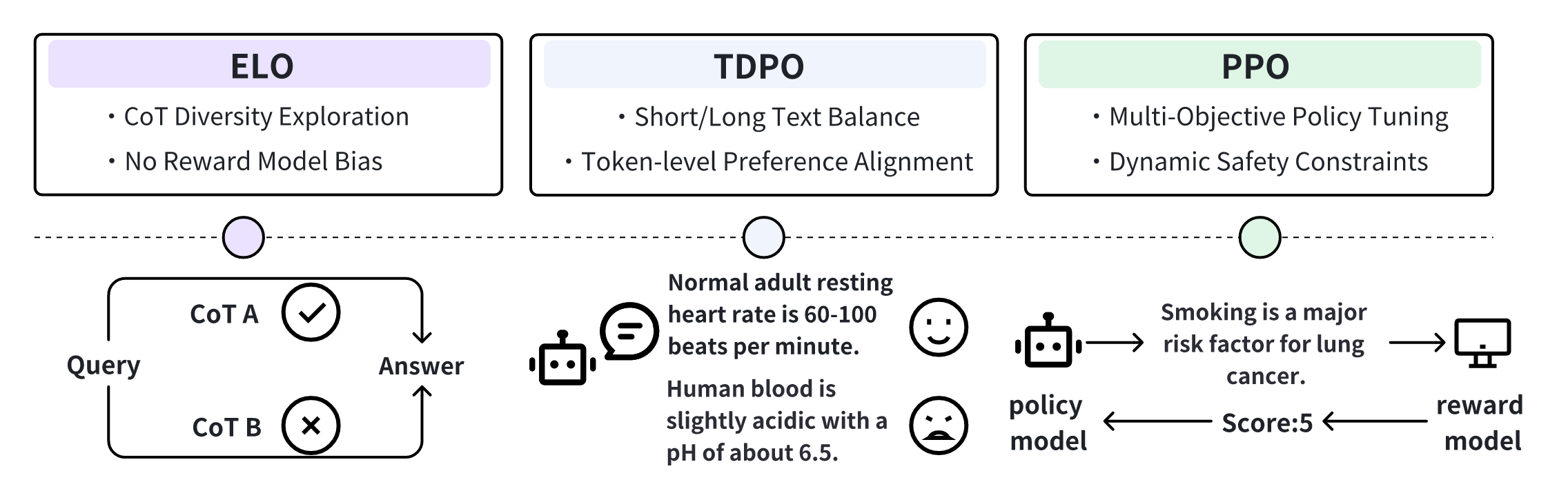}
	\caption{The pipeline of our three-staged reinforcement learning.}
	\label{fig:rl}
\end{figure}
As shown in Figure~\ref{fig:rl}, we apply a three-staged reinforcement learning process to further boost the reasoning ability of our model. This structured approach ensures that the model not only generates high-quality outputs but also aligns with user preferences and maintains logical coherence across diverse tasks. Each stage—Exploratory Log-likelihood Optimization (ELO), Token-Level Direct Preference Optimization (TDPO)~\cite{zeng2024token}, and Proximal Policy Optimization (PPO)~\cite{schulman2017proximal}—plays a distinct role in refining the model's capabilities. \\
\noindent\textbf{Exploratory Log-likelihood Optimization} 
The ELO phase is designed to enhance the model's ability to generate diverse and high-quality chain-of-thought (CoT) reasoning paths. Unlike traditional reinforcement learning methods that depend on a reward model to guide optimization, ELO directly optimizes the likelihood of generating coherent and logical reasoning paths. This is achieved by maximizing the probability of high-quality outputs given the input, without relying on external reward signals.
For user queries $Q$, ELO aims to maximize the log-likelihood of the ground truth answer $A$ by optimizing the distribution of CoT generated by model $M$:
\begin{equation}
	\begin{aligned}
		L_{ELO}=-\mathbb{E}_Q \log \pi_M (A|Q)=-\mathbb{E}_Q\log \mathbb{E}_{\text{CoT} \sim M}\pi_M(A|Q,\text{CoT}).
	\end{aligned}
\end{equation}
According to Jensen's inequality, we can derive the variational upper bound of $L_{ELO}$:
\begin{equation}
	\begin{aligned}
		L_{ELO} \leq L_{upper}=-\mathbb{E}_Q \mathbb{E}_{\text{CoT} \sim M}\log\pi_M(A|Q,\text{CoT}).
	\end{aligned}
\end{equation}
We can derive the gradient of the $L_{upper}$ as
\begin{equation}
	\begin{aligned}
		\nabla L_{upper}=-\mathbb{E}_Q \mathbb{E}_{\text{CoT} \sim M}(\log\pi_M(A|Q,\text{CoT})-b(Q,A))\nabla \log \pi_M(\text{CoT}|Q),
	\end{aligned}
\end{equation}
which can be directly optimized based on the log-likelihood of the ground truth answer.
This approach eliminates potential biases introduced by reward models, ensuring more stable and logical reasoning generation.
\\
\noindent\textbf{Token-Level Direct Preference Optimization} 
Building on the ELO-trained model, the TDPO phase refines the model using preference pair data. Traditional DPO methods often struggle with length-dependent constraints due to the nature of the KL divergence term, which enforces alignment between the generated outputs and a reference model. In DPO, the KL divergence tends to impose stronger constraints on shorter sequences while being less effective for longer ones. This imbalance can lead to suboptimal performance, especially in tasks requiring long-form reasoning or detailed explanations. 
TDPO addresses this limitation by introducing token-level optimization. Instead of applying constraints at the sequence level, TDPO operates at the token level, ensuring that the model maintains alignment with user preferences across both short and long sequences.  \\
\noindent\textbf{Proximal Policy optimization}
In the final stage, PPO is employed to further refine the model's generation strategy. PPO leverages the improvements achieved during the ELO and TDPO phases, combining them with feedback from a reward model to fine-tune the model's policy. The reward model provides real-time feedback on the quality of generated outputs, ensuring that the model aligns with user preferences and performs well across a variety of tasks.

\section{Evaluations}

\subsection{Benchmarks}

Our evaluation covers mainstream open-source datasets and private datasets.
We categorize medical capabilities into three levels: medical fundamentals, medical examinations, and medical practice.

The medical fundamentals level includes tasks such as MedNLI \citep{romanov2018mednli} and MedCalc \citep{khandekar2024medcalcbench}. 
We also include some medical related subjects of MMLU \cite{hendrycks2020measuring}.
\begin{itemize}
	\item MedNLI is a natural language inference task, that is, to determine whether two medical statements are inclusive, mutually exclusive, or irrelevant.
	\item MedCalc is a medical calculation dataset. The model is asked to compute a clinical value based on a patient note.
	\item MMLU is a well known benchmark for large language model evaluation, which covering 57 tasks.
\end{itemize}
Medical examinations focus on various real medical exams, including USMLE \citep{jin2021disease}, CMExam \citep{liu2024cmexam}, MediQ \citep{li2024mediq}, MedBullets \citep{chen2024medbullets}, Pubmedqa \citep{jin2019pubmedqa}, ReDis-QA \citep{wang2024redisqa}, as well as intermediate-level physician exam questions collected from the Chinese Internet, covering pediatrics (Erke), internal medicine (Neike), and general practice (Quanke). 
\begin{itemize}
	\item USMLE is a subset of MedQA \citep{jin2021disease}, whcih is collected form the professional medical board exams of US.
	
	\item MediQ simulates an interactive conversation between a patient and an expert. This benchmark evaluates how well the participants' expert modules can handle realistic patient queries by either asking relevant questions or making final decisions based on the conversation history.
	
	\item MedBullets contains 308 questions of the USMLE Step 2 \& 3 type. The questions are mainly collected from Twitter after 2022.
	\item Pubmedqa is a biomedical-related question-answer dataset constructed from the abstract section of PubMed. Given a medical research question, you are required to answer it with yes/no/maybe based on the provided abstract.
	
	\item ReDis-QA is constructed around the diagnosis of rare diseases, covering 205 types of rare diseases. It is used to evaluate the performance of large language models in rare disease diagnosis.
	\item We created Erke, Neike and Quanke based on intermediate-level physician exam questions collected from the Chinese Internet.
	
\end{itemize}

The medical practice tier focuses on real consultations, including CMBClin \citep{wang2023cmb}, ClinicalBench \citep{yan2024clinicalbench}, RareArena \citep{RareArena}, RareBench \citep{chen2024rarebench} and NEJMQA.
\begin{itemize}
	\item CMBExam and CMBClin cover all levels of clinical medical specialties and comprehensively evaluate the model's medical knowledge and clinical consultation abilities.
	\item ClinicalBench contains clinical diagnosis evaluations covering 150 diseases in 24 departments based on real cases. There are a total of 1,500 samples. By simulating the complete medical treatment process, 8 tasks are set up to evaluate the model from the two dimensions of tasks and departments.
	\item RareArena is a dataset of nearly 50,000 rare disease diagnoses extracted from case summaries in PubMed Central, covering 4,597 rare disease types.
	It is divided into two settings: RDS (rare disease screening), where the input does not include specific examination results; and RDC (rare disease confirmation), where the input includes specific examinations.
	\item RareBench covers patient electronic health records (EHRs), symptoms, and confirmed diagnosis information, involving numerous rare diseases and some common diseases. 
	\item We created  NEJMQA dataset using case challenges from The New England Journal of Medicine\footnote{https://www.nejm.org/} (NEJM) to construct diagnostic questions, including nearly 200 patients.
\end{itemize}

For each benchmark, we randomly select at most 1000 samples in our evaluation.

\subsection{Response and Scoring}

The prompts used to obtain responses for different benchmarks are listed in Appendix \ref{evalpromptsappendix} Table \ref{responseprompt}.

The generated responses are all free text. 
We use open-source Qwen2.5-72B-Instruct \citep{qwen2025qwen25technicalreport} to perform post-processing or scoring on the responses.

For multiple-choice questions, we first extract the selected option labels and then compare them with the correct answers to determine the score.
For other benchmarks, we acquire the score based on the prompts in Appendix \ref{evalpromptsappendix} Table \ref{scoreprompt} directly.
The scores for each benchmark will be normalized.

\begin{table}[]
	\begin{center}
		\begin{tabular}{lccccc}
			\hline
			\multicolumn{1}{c}{\textbf{}} & 
			\multicolumn{1}{l}{\makecell[c]{Baichuan\\-M1-14B\\-Instruct} }& 
			\multicolumn{1}{l}{\makecell[c]{Qwen2.5\\-14B\\-Instruct} }& 
			\multicolumn{1}{l}{\makecell[c]{Qwen2.5\\-72B\\-Instruct} }& 
			\multicolumn{1}{l}{\makecell[c]{claude\\-3.5\\-sonnet} }& 
			\multicolumn{1}{l}{\makecell[c]{gpt-4o}} \\
			\hline
			\textbf{Average}              & 72.23              & 65.39 & 70.51 & 74.85 & 75.00 \\
			\hline
			
			\makecell[l]{CMBClin}              & 77.40                        & 71.51 & 75.36 & 78.37 & 75.36 \\
			\makecell[l]{ClinicalBench-Diagnosis}  & {\color[HTML]{212529} 70.90} & 68.85 & 72.23 & 75.00 & 73.05 \\
			\makecell[l]{ClinicalBench-Department}   & {\color[HTML]{212529} 70.05} & 68.83 & 70.53 & 65.58 & 69.38 \\
			\makecell[l]{ClinicalBench-Treatment} & 56.38                        & 55.03 & 57.30 & 64.03 & 59.35 \\
			\makecell[l]{RareArena-rdc}       & 81.80                        & 66.40 & 76.20 & 89.60 & 88.40 \\
			\makecell[l]{RareArena-rds}       & 54.00                        & 42.60 & 49.80 & 59.80 & 57.20 \\
			\makecell[l]{RareBench}            & 59.60                        & 52.80 & 60.60 & 65.30 & 62.80 \\
			\makecell[l]{NEJMQA}               & 49.75                        & 45.69 & 50.76 & 69.54 & 54.31 \\
			\hline
			
			\makecell[l]{CMExam}               & 80.10                        & 77.70 & 82.70 & 77.50 & 78.00 \\
			\makecell[l]{Erke}                 & 78.48                        & 74.68 & 84.81 & 76.58 & 78.48 \\
			\makecell[l]{Neike}                & 83.42                        & 86.10 & 87.17 & 87.70 & 83.42 \\
			\makecell[l]{Quanke}               & 87.07                        & 88.44 & 88.44 & 81.63 & 84.35 \\
			\makecell[l]{USMLE}                & 78.00                        & 67.20 & 76.70 & 85.90 & 87.10 \\
			\makecell[l]{MedBullets}           & 66.88                        & 54.22 & 64.29 & 72.40 & 75.97 \\
			\makecell[l]{MediQ}                & 83.40                        & 66.80 & 79.90 & 88.80 & 90.20 \\
			
			\makecell[l]{Pubmedqa}             & 75.20                        & 76.40 & 75.60 & 77.00 & 77.60 \\
			\makecell[l]{ReDis-QA}             & 74.50                         & 69.70 & 75.00 & 83.20 & 82.80 \\
			\hline
			
			\makecell[l]{MedNLI-Dis}          & 80.40                        & 68.90 & 74.90 & 58.30 & 79.80 \\
			\makecell[l]{MedCalc}              & 56.00                        & 31.40 & 37.90 & 52.60 & 49.00 \\
			\makecell[l]{MMLU-anatomy}         & 80.00                           & 67.41 & 71.11 & 86.67 & 91.11 \\
			\makecell[l]{MMLU-virology}        & 54.82                        & 56.02 & 53.01 & 54.22 & 57.23 \\
			\makecell[l]{MMLU-genetics}        & 91.00                           & 82.00 & 87.00 & 97.00 & 95.00 \\
			
			\hline
			
		\end{tabular}
		\caption{Results of Baichuan-M1-14B-Instruct, compared with strong baselines. }
		\label{evalresults}
	\end{center}
\end{table}

\subsection{Results}

As shown in Table \ref{evalresults}, Baichuan-M1-14B-Instruct surpasses Qwen2.5-72B-Instruct, which is one of the strongest open-source models, in the area of medicine. 
Although Baichuan-M1-14B-Instruct is still lagging behind Claude-3.5-Sonnet\footnote{claude-3.5-sonnet-20241022} and GPT-4o\footnote{gpt-4o-2024-05-13}, which are the most representative proprietary models, the gap is no longer significant.

We show some examples in Appendix \ref{evalpromptsappendix} Table \ref{examplesofbc}.

In addition, we also show the code and math ability of Baichuan-M1-14B-Base, compared with the Qwen2.5 series models~\citep{qwen2025qwen25technicalreport}. 
We tested the performance of code generation based on the EvalPlus framework~\citep{liu2024your} and Bigcodebench~\citep{zhuo2024bigcodebench}. 
We use MATH~\citep{hendrycks2021measuring} and CMATH~\citep{wei2023cmath} to evaluate the math ability.

\begin{table}[]
	\begin{center}
		\begin{tabular}{lcccc}
			\hline
			Model & 
			\multicolumn{1}{l}{\makecell[c]{Baichuan-M1\\-14B-Base}} & 
			\multicolumn{1}{l}{Qwen2.5-14B} & 
			\multicolumn{1}{l}{Qwen2.5-32B} & 
			\multicolumn{1}{l}{Qwen2.5-72B} \\
			\hline
			MBPP         & 74.0 & 72.8 & 83.3 & 86.5 \\
			MBPP+        & 63.0 & 63.2 & 69.0 & 70.1 \\
			HumanEval    & 60.4 & 56.7 & 58.5 & 59.1 \\
			HumanEval+   & 53.7 & 51.2 & 54.3 & 54.9 \\
			Bigcodebench & 48.7  & 46.8  & 47.2  & 50.3  \\
			\hline
			MATH         & 46.0  & 45.4  & 50.6  & 48.2  \\
			CMATH        & 88.3  & 88.7  & 86.8  & -  \\
			\hline
		\end{tabular}
		\caption{Code and math ability of Baichuan-M1-14B-Base.}
		\label{mathcode}
	\end{center}
\end{table}

\section{Discussion and Conclusion}
\subsection{Continue Pre-training or Training from Scratch?}
Most previous works applying large language models to specialized domains involve additional training on an off-the-shelf base model \cite{dou2024sailor,singhal2025toward,zhang2023xuanyuan}. This approach is considered the most effective for improving vertical capabilities while maintaining general performance. However, in our preliminary experiments, we found that continuing training on well-trained base models, especially those that have undergone annealing, makes it difficult to improve vertical capabilities without sacrificing general capabilities. 

In the medical domain, the language used and the knowledge involved are significantly different from those in general domains. Therefore, the approach of continuing training by simply increasing the proportion of domain-specific data or adjusting the learning rate may be ineffective, if not impossible, in truly improving medical capabilities, compared to post-training or training from scratch. More future work should focus on this interesting and important task.

\subsection{Conclusion and Future Work}
In conclusion, the introduction of Baichuan-M1 marks a significant advancement in the application of large language models (LLMs) to the medical domain. By training from scratch with a dedicated focus on medical expertise, Baichuan-M1 overcomes the limitations of traditional approaches that rely on fine-tuning general-purpose models. This approach has proven to be particularly effective in handling the complexities of medical knowledge, enabling the model to perform robustly across a range of medical applications, from diagnostics to treatment recommendations.

The model’s strength lies not only in its ability to integrate vast amounts of medical literature, clinical data, and expert-curated sources but also in its sophisticated training methodologies, including the use of synthetic data and a finely tuned pre-training process. The balance between general capabilities and specialized medical knowledge ensures that Baichuan-M1 excels in both common language tasks and highly specialized medical queries.

Furthermore, the open-sourcing of Baichuan-M1-14B provides an important resource for the broader research community, facilitating further exploration and refinement of medical LLMs. While there is still room for improvement, especially in the areas of rare disease diagnosis and real-world clinical consultation, the results demonstrate that Baichuan-M1 represents a promising leap forward in developing LLMs that can meet the precise demands of the medical field. The model’s continued evolution will likely contribute to the enhancement of AI-driven medical decision-making, offering both improved accuracy and greater potential for advancing healthcare technologies.

\newpage
\bibliography{baichuan}

\begin{thebibliography}{81}
\providecommand{\natexlab}[1]{#1}
\providecommand{\url}[1]{\texttt{#1}}
\expandafter\ifx\csname urlstyle\endcsname\relax
  \providecommand{\doi}[1]{doi: #1}\else
  \providecommand{\doi}{doi: \begingroup \urlstyle{rm}\Url}\fi

\bibitem[Rar(2024)]{RareArena}
Rarearena.
\newblock \url{https://github.com/zhao-zy15/RareArena/}, 2024.

\bibitem[Abdin et~al.(2024)Abdin, Aneja, Behl, Bubeck, Eldan, Gunasekar,
  Harrison, Hewett, Javaheripi, Kauffmann, Lee, Lee, Li, Liu, Mendes, Nguyen,
  Price, de~Rosa, Saarikivi, Salim, Shah, Wang, Ward, Wu, Yu, Zhang, and
  Zhang]{abdin2024phi4technicalreport}
Marah Abdin, Jyoti Aneja, Harkirat Behl, Sébastien Bubeck, Ronen Eldan, Suriya
  Gunasekar, Michael Harrison, Russell~J. Hewett, Mojan Javaheripi, Piero
  Kauffmann, James~R. Lee, Yin~Tat Lee, Yuanzhi Li, Weishung Liu, Caio C.~T.
  Mendes, Anh Nguyen, Eric Price, Gustavo de~Rosa, Olli Saarikivi, Adil Salim,
  Shital Shah, Xin Wang, Rachel Ward, Yue Wu, Dingli Yu, Cyril Zhang, and
  Yi~Zhang.
\newblock Phi-4 technical report, 2024.
\newblock URL \url{https://arxiv.org/abs/2412.08905}.

\bibitem[Adeshola \& Adepoju(2024)Adeshola and
  Adepoju]{adeshola2024opportunities}
Ibrahim Adeshola and Adeola~Praise Adepoju.
\newblock The opportunities and challenges of chatgpt in education.
\newblock \emph{Interactive Learning Environments}, 32\penalty0 (10):\penalty0
  6159--6172, 2024.

\bibitem[Austin et~al.(2021)Austin, Odena, Nye, Bosma, Michalewski, Dohan,
  Jiang, Cai, Terry, Le, and Sutton]{austin2021programsynthesislargelanguage}
Jacob Austin, Augustus Odena, Maxwell Nye, Maarten Bosma, Henryk Michalewski,
  David Dohan, Ellen Jiang, Carrie Cai, Michael Terry, Quoc Le, and Charles
  Sutton.
\newblock Program synthesis with large language models, 2021.
\newblock URL \url{https://arxiv.org/abs/2108.07732}.

\bibitem[Bai et~al.(2023)Bai, Bai, Chu, Cui, Dang, Deng, Fan, Ge, Han, Huang,
  et~al.]{bai2023qwen}
Jinze Bai, Shuai Bai, Yunfei Chu, Zeyu Cui, Kai Dang, Xiaodong Deng, Yang Fan,
  Wenbin Ge, Yu~Han, Fei Huang, et~al.
\newblock Qwen technical report.
\newblock \emph{arXiv preprint arXiv:2309.16609}, 2023.

\bibitem[Beck et~al.(2024)Beck, P{\"o}ppel, Spanring, Auer, Prudnikova, Kopp,
  Klambauer, Brandstetter, and Hochreiter]{beck2024xlstm}
Maximilian Beck, Korbinian P{\"o}ppel, Markus Spanring, Andreas Auer,
  Oleksandra Prudnikova, Michael Kopp, G{\"u}nter Klambauer, Johannes
  Brandstetter, and Sepp Hochreiter.
\newblock xlstm: Extended long short-term memory.
\newblock \emph{arXiv preprint arXiv:2405.04517}, 2024.

\bibitem[Bisk et~al.(2020)Bisk, Zellers, Gao, Choi, et~al.]{bisk2020piqa}
Yonatan Bisk, Rowan Zellers, Jianfeng Gao, Yejin Choi, et~al.
\newblock Piqa: Reasoning about physical commonsense in natural language.
\newblock In \emph{Proceedings of the AAAI conference on artificial
  intelligence}, volume~34, pp.\  7432--7439, 2020.

\bibitem[Chen et~al.(2024{\natexlab{a}})Chen, Fang, Singla, and
  Dredze]{chen2024medbullets}
Hanjie Chen, Zhouxiang Fang, Yash Singla, and Mark Dredze.
\newblock Benchmarking large language models on answering and explaining
  challenging medical questions.
\newblock \emph{arXiv preprint arXiv:2402.18060}, 2024{\natexlab{a}}.

\bibitem[Chen et~al.(2024{\natexlab{b}})Chen, Waheed, Li, Wang, Wang, Raj, and
  Abdin]{chen2024diversitysyntheticdataimpact}
Hao Chen, Abdul Waheed, Xiang Li, Yidong Wang, Jindong Wang, Bhiksha Raj, and
  Marah~I. Abdin.
\newblock On the diversity of synthetic data and its impact on training large
  language models, 2024{\natexlab{b}}.
\newblock URL \url{https://arxiv.org/abs/2410.15226}.

\bibitem[Chen et~al.(2024{\natexlab{c}})Chen, Cai, Ji, Wang, Liu, Wang, Hou,
  and Wang]{chen2024huatuogpt}
Junying Chen, Zhenyang Cai, Ke~Ji, Xidong Wang, Wanlong Liu, Rongsheng Wang,
  Jianye Hou, and Benyou Wang.
\newblock Huatuogpt-o1, towards medical complex reasoning with llms.
\newblock \emph{arXiv preprint arXiv:2412.18925}, 2024{\natexlab{c}}.

\bibitem[Chen et~al.(2024{\natexlab{d}})Chen, Wang, Palangi, Han, Ebrahimi, Le,
  Perot, Mishra, Bansal, Lee, and Pfister]{chen2024reversethinkingmakesllms}
Justin Chih-Yao Chen, Zifeng Wang, Hamid Palangi, Rujun Han, Sayna Ebrahimi,
  Long Le, Vincent Perot, Swaroop Mishra, Mohit Bansal, Chen-Yu Lee, and Tomas
  Pfister.
\newblock Reverse thinking makes llms stronger reasoners, 2024{\natexlab{d}}.
\newblock URL \url{https://arxiv.org/abs/2411.19865}.

\bibitem[Chen et~al.(2024{\natexlab{e}})Chen, Mao, Guo, Wang, Zhang, and
  Chen]{chen2024rarebench}
Xuanzhong Chen, Xiaohao Mao, Qihan Guo, Lun Wang, Shuyang Zhang, and Ting Chen.
\newblock Rarebench: Can llms serve as rare diseases specialists?
\newblock In \emph{Proceedings of the 30th ACM SIGKDD Conference on Knowledge
  Discovery and Data Mining}, pp.\  4850--4861, 2024{\natexlab{e}}.

\bibitem[Clark et~al.(2018)Clark, Cowhey, Etzioni, Khot, Sabharwal, Schoenick,
  and Tafjord]{clark2018think}
Peter Clark, Isaac Cowhey, Oren Etzioni, Tushar Khot, Ashish Sabharwal, Carissa
  Schoenick, and Oyvind Tafjord.
\newblock Think you have solved question answering? try arc, the ai2 reasoning
  challenge.
\newblock \emph{arXiv preprint arXiv:1803.05457}, 2018.

\bibitem[DeepSeek-AI et~al.(2024)DeepSeek-AI, Liu, Feng, Xue, Wang, Wu, Lu,
  Zhao, Deng, Zhang, Ruan, Dai, Guo, Yang, Chen, Ji, Li, Lin, Dai, Luo, Hao,
  Chen, Li, Zhang, Bao, Xu, Wang, Zhang, Ding, Xin, Gao, Li, Qu, Cai, Liang,
  Guo, Ni, Li, Wang, Chen, Chen, Yuan, Qiu, Li, Song, Dong, Hu, Gao, Guan,
  Huang, Yu, Wang, Zhang, Xu, Xia, Zhao, Wang, Zhang, Li, Wang, Zhang, Zhang,
  Tang, Li, Tian, Huang, Wang, Zhang, Wang, Zhu, Chen, Du, Chen, Jin, Ge,
  Zhang, Pan, Wang, Xu, Zhang, Chen, Li, Lu, Zhou, Chen, Wu, Ye, Ye, Ma, Wang,
  Zhou, Yu, Zhou, Pan, Wang, Yun, Pei, Sun, Xiao, Zeng, Zhao, An, Liu, Liang,
  Gao, Yu, Zhang, Li, Jin, Wang, Bi, Liu, Wang, Shen, Chen, Zhang, Chen, Nie,
  Sun, Wang, Cheng, Liu, Xie, Liu, Yu, Song, Shan, Zhou, Yang, Li, Su, Lin, Li,
  Wang, Wei, Zhu, Zhang, Xu, Xu, Huang, Li, Zhao, Sun, Li, Wang, Yu, Zheng,
  Zhang, Shi, Xiong, He, Tang, Piao, Wang, Tan, Ma, Liu, Guo, Wu, Ou, Zhu,
  Wang, Gong, Zou, He, Zha, Xiong, Ma, Yan, Luo, You, Liu, Zhou, Wu, Ren, Ren,
  Sha, Fu, Xu, Huang, Zhang, Xie, Zhang, Hao, Gou, Ma, Yan, Shao, Xu, Wu,
  Zhang, Li, Gu, Zhu, Liu, Li, Xie, Song, Gao, and
  Pan]{deepseekai2024deepseekv3technicalreport}
DeepSeek-AI, Aixin Liu, Bei Feng, Bing Xue, Bingxuan Wang, Bochao Wu, Chengda
  Lu, Chenggang Zhao, Chengqi Deng, Chenyu Zhang, Chong Ruan, Damai Dai, Daya
  Guo, Dejian Yang, Deli Chen, Dongjie Ji, Erhang Li, Fangyun Lin, Fucong Dai,
  Fuli Luo, Guangbo Hao, Guanting Chen, Guowei Li, H.~Zhang, Han Bao, Hanwei
  Xu, Haocheng Wang, Haowei Zhang, Honghui Ding, Huajian Xin, Huazuo Gao, Hui
  Li, Hui Qu, J.~L. Cai, Jian Liang, Jianzhong Guo, Jiaqi Ni, Jiashi Li, Jiawei
  Wang, Jin Chen, Jingchang Chen, Jingyang Yuan, Junjie Qiu, Junlong Li,
  Junxiao Song, Kai Dong, Kai Hu, Kaige Gao, Kang Guan, Kexin Huang, Kuai Yu,
  Lean Wang, Lecong Zhang, Lei Xu, Leyi Xia, Liang Zhao, Litong Wang, Liyue
  Zhang, Meng Li, Miaojun Wang, Mingchuan Zhang, Minghua Zhang, Minghui Tang,
  Mingming Li, Ning Tian, Panpan Huang, Peiyi Wang, Peng Zhang, Qiancheng Wang,
  Qihao Zhu, Qinyu Chen, Qiushi Du, R.~J. Chen, R.~L. Jin, Ruiqi Ge, Ruisong
  Zhang, Ruizhe Pan, Runji Wang, Runxin Xu, Ruoyu Zhang, Ruyi Chen, S.~S. Li,
  Shanghao Lu, Shangyan Zhou, Shanhuang Chen, Shaoqing Wu, Shengfeng Ye,
  Shengfeng Ye, Shirong Ma, Shiyu Wang, Shuang Zhou, Shuiping Yu, Shunfeng
  Zhou, Shuting Pan, T.~Wang, Tao Yun, Tian Pei, Tianyu Sun, W.~L. Xiao,
  Wangding Zeng, Wanjia Zhao, Wei An, Wen Liu, Wenfeng Liang, Wenjun Gao,
  Wenqin Yu, Wentao Zhang, X.~Q. Li, Xiangyue Jin, Xianzu Wang, Xiao Bi,
  Xiaodong Liu, Xiaohan Wang, Xiaojin Shen, Xiaokang Chen, Xiaokang Zhang,
  Xiaosha Chen, Xiaotao Nie, Xiaowen Sun, Xiaoxiang Wang, Xin Cheng, Xin Liu,
  Xin Xie, Xingchao Liu, Xingkai Yu, Xinnan Song, Xinxia Shan, Xinyi Zhou,
  Xinyu Yang, Xinyuan Li, Xuecheng Su, Xuheng Lin, Y.~K. Li, Y.~Q. Wang, Y.~X.
  Wei, Y.~X. Zhu, Yang Zhang, Yanhong Xu, Yanhong Xu, Yanping Huang, Yao Li,
  Yao Zhao, Yaofeng Sun, Yaohui Li, Yaohui Wang, Yi~Yu, Yi~Zheng, Yichao Zhang,
  Yifan Shi, Yiliang Xiong, Ying He, Ying Tang, Yishi Piao, Yisong Wang, Yixuan
  Tan, Yiyang Ma, Yiyuan Liu, Yongqiang Guo, Yu~Wu, Yuan Ou, Yuchen Zhu, Yuduan
  Wang, Yue Gong, Yuheng Zou, Yujia He, Yukun Zha, Yunfan Xiong, Yunxian Ma,
  Yuting Yan, Yuxiang Luo, Yuxiang You, Yuxuan Liu, Yuyang Zhou, Z.~F. Wu,
  Z.~Z. Ren, Zehui Ren, Zhangli Sha, Zhe Fu, Zhean Xu, Zhen Huang, Zhen Zhang,
  Zhenda Xie, Zhengyan Zhang, Zhewen Hao, Zhibin Gou, Zhicheng Ma, Zhigang Yan,
  Zhihong Shao, Zhipeng Xu, Zhiyu Wu, Zhongyu Zhang, Zhuoshu Li, Zihui Gu,
  Zijia Zhu, Zijun Liu, Zilin Li, Ziwei Xie, Ziyang Song, Ziyi Gao, and Zizheng
  Pan.
\newblock Deepseek-v3 technical report, 2024.
\newblock URL \url{https://arxiv.org/abs/2412.19437}.

\bibitem[Dou et~al.(2024)Dou, Liu, Zeng, Guo, Zhou, Lu, and Lin]{dou2024sailor}
Longxu Dou, Qian Liu, Guangtao Zeng, Jia Guo, Jiahui Zhou, Wei Lu, and Min Lin.
\newblock Sailor: Open language models for south-east asia.
\newblock \emph{arXiv preprint arXiv:2404.03608}, 2024.

\bibitem[Fu et~al.(2022)Fu, Dao, Saab, Thomas, Rudra, and R{\'e}]{fu2022hungry}
Daniel~Y Fu, Tri Dao, Khaled~K Saab, Armin~W Thomas, Atri Rudra, and
  Christopher R{\'e}.
\newblock Hungry hungry hippos: Towards language modeling with state space
  models.
\newblock \emph{arXiv preprint arXiv:2212.14052}, 2022.

\bibitem[Guo et~al.(2025)Guo, Yang, Zhang, Song, Zhang, Xu, Zhu, Ma, Wang, Bi,
  et~al.]{guo2025deepseek}
Daya Guo, Dejian Yang, Haowei Zhang, Junxiao Song, Ruoyu Zhang, Runxin Xu,
  Qihao Zhu, Shirong Ma, Peiyi Wang, Xiao Bi, et~al.
\newblock Deepseek-r1: Incentivizing reasoning capability in llms via
  reinforcement learning.
\newblock \emph{arXiv preprint arXiv:2501.12948}, 2025.

\bibitem[Hendrycks et~al.(2020)Hendrycks, Burns, Basart, Zou, Mazeika, Song,
  and Steinhardt]{hendrycks2020measuring}
Dan Hendrycks, Collin Burns, Steven Basart, Andy Zou, Mantas Mazeika, Dawn
  Song, and Jacob Steinhardt.
\newblock Measuring massive multitask language understanding.
\newblock \emph{arXiv preprint arXiv:2009.03300}, 2020.

\bibitem[Hendrycks et~al.(2021)Hendrycks, Burns, Basart, Zou, Mazeika, Song,
  and Steinhardt]{hendrycks2021measuring}
Dan Hendrycks, Collin Burns, Steven Basart, Andy Zou, Mantas Mazeika, Dawn
  Song, and Jacob Steinhardt.
\newblock Measuring massive multitask language understanding, 2021.

\bibitem[Hu et~al.(2024)Hu, Tu, Han, He, Cui, Long, Zheng, Fang, Huang, Zhao,
  et~al.]{hu2024minicpm}
Shengding Hu, Yuge Tu, Xu~Han, Chaoqun He, Ganqu Cui, Xiang Long, Zhi Zheng,
  Yewei Fang, Yuxiang Huang, Weilin Zhao, et~al.
\newblock Minicpm: Unveiling the potential of small language models with
  scalable training strategies.
\newblock \emph{arXiv preprint arXiv:2404.06395}, 2024.

\bibitem[Huang et~al.(2023)Huang, Bai, Zhu, Zhang, Zhang, Su, Liu, Lv, Zhang,
  Lei, Fu, Sun, and He]{huang2023ceval}
Yuzhen Huang, Yuzhuo Bai, Zhihao Zhu, Junlei Zhang, Jinghan Zhang, Tangjun Su,
  Junteng Liu, Chuancheng Lv, Yikai Zhang, Jiayi Lei, Yao Fu, Maosong Sun, and
  Junxian He.
\newblock C-eval: A multi-level multi-discipline chinese evaluation suite for
  foundation models.
\newblock In \emph{Advances in Neural Information Processing Systems}, 2023.

\bibitem[Hui et~al.(2024)Hui, Yang, Cui, Yang, Liu, Zhang, Liu, Zhang, Yu, Lu,
  Dang, Fan, Zhang, Yang, Men, Huang, Zheng, Miao, Quan, Feng, Ren, Ren, Zhou,
  and Lin]{qwen2.5_coder}
Binyuan Hui, Jian Yang, Zeyu Cui, Jiaxi Yang, Dayiheng Liu, Lei Zhang, Tianyu
  Liu, Jiajun Zhang, Bowen Yu, Keming Lu, Kai Dang, Yang Fan, Yichang Zhang,
  An~Yang, Rui Men, Fei Huang, Bo~Zheng, Yibo Miao, Shanghaoran Quan, Yunlong
  Feng, Xingzhang Ren, Xuancheng Ren, Jingren Zhou, and Junyang Lin.
\newblock Qwen2.5-coder technical report.
\newblock \emph{arXiv preprint arXiv:2409.12186}, 2024.
\newblock URL \url{https://arxiv.org/abs/2409.12186}.

\bibitem[Jin et~al.(2021)Jin, Pan, Oufattole, Weng, Fang, and
  Szolovits]{jin2021disease}
Di~Jin, Eileen Pan, Nassim Oufattole, Wei-Hung Weng, Hanyi Fang, and Peter
  Szolovits.
\newblock What disease does this patient have? a large-scale open domain
  question answering dataset from medical exams.
\newblock \emph{Applied Sciences}, 11\penalty0 (14):\penalty0 6421, 2021.

\bibitem[Jin et~al.(2019)Jin, Dhingra, Liu, Cohen, and Lu]{jin2019pubmedqa}
Qiao Jin, Bhuwan Dhingra, Zhengping Liu, William~W Cohen, and Xinghua Lu.
\newblock Pubmedqa: A dataset for biomedical research question answering.
\newblock \emph{arXiv preprint arXiv:1909.06146}, 2019.

\bibitem[Johnson et~al.(2016)Johnson, Pollard, Shen, Lehman, Feng, Ghassemi,
  Moody, Szolovits, Anthony~Celi, and Mark]{johnson2016mimic}
Alistair~EW Johnson, Tom~J Pollard, Lu~Shen, Li-wei~H Lehman, Mengling Feng,
  Mohammad Ghassemi, Benjamin Moody, Peter Szolovits, Leo Anthony~Celi, and
  Roger~G Mark.
\newblock Mimic-iii, a freely accessible critical care database.
\newblock \emph{Scientific data}, 3\penalty0 (1):\penalty0 1--9, 2016.

\bibitem[Johnson et~al.(2023)Johnson, Bulgarelli, Shen, Gayles, Shammout,
  Horng, Pollard, Hao, Moody, Gow, et~al.]{johnson2023mimic}
Alistair~EW Johnson, Lucas Bulgarelli, Lu~Shen, Alvin Gayles, Ayad Shammout,
  Steven Horng, Tom~J Pollard, Sicheng Hao, Benjamin Moody, Brian Gow, et~al.
\newblock Mimic-iv, a freely accessible electronic health record dataset.
\newblock \emph{Scientific data}, 10\penalty0 (1):\penalty0 1, 2023.

\bibitem[Khandekar et~al.(2024)Khandekar, Jin, Xiong, Dunn, Applebaum, Anwar,
  Sarfo-Gyamfi, Safranek, Anwar, Zhang, Gilson, Singer, Dave, Taylor, Zhang,
  Chen, and Lu]{khandekar2024medcalcbench}
Nikhil Khandekar, Qiao Jin, Guangzhi Xiong, Soren Dunn, Serina~S Applebaum,
  Zain Anwar, Maame Sarfo-Gyamfi, Conrad~W Safranek, Abid~A Anwar, Andrew
  Zhang, Aidan Gilson, Maxwell~B Singer, Amisha Dave, Andrew Taylor, Aidong
  Zhang, Qingyu Chen, and Zhiyong Lu.
\newblock Medcalc-bench: Evaluating large language models for medical
  calculations, 2024.

\bibitem[Kudo(2018)]{kudo2018sentencepiece}
T~Kudo.
\newblock Sentencepiece: A simple and language independent subword tokenizer
  and detokenizer for neural text processing.
\newblock \emph{arXiv preprint arXiv:1808.06226}, 2018.

\bibitem[Li et~al.(2024{\natexlab{a}})Li, Zhang, Koto, Yang, Zhao, Gong, Duan,
  and Baldwin]{li2024cmmlu}
Haonan Li, Yixuan Zhang, Fajri Koto, Yifei Yang, Hai Zhao, Yeyun Gong, Nan
  Duan, and Timothy Baldwin.
\newblock Cmmlu: Measuring massive multitask language understanding in chinese,
  2024{\natexlab{a}}.

\bibitem[Li et~al.(2024{\natexlab{b}})Li, Balachandran, Feng, Ilgen, Pierson,
  Koh, and Tsvetkov]{li2024mediq}
Shuyue~Stella Li, Vidhisha Balachandran, Shangbin Feng, Jonathan Ilgen, Emma
  Pierson, Pang~Wei Koh, and Yulia Tsvetkov.
\newblock Mediq: Question-asking llms for adaptive and reliable medical
  reasoning.
\newblock \emph{arXiv preprint arXiv:2406.00922}, 2024{\natexlab{b}}.

\bibitem[Liu et~al.(2024{\natexlab{a}})Liu, Xia, Wang, and Zhang]{liu2024your}
Jiawei Liu, Chunqiu~Steven Xia, Yuyao Wang, and Lingming Zhang.
\newblock Is your code generated by chatgpt really correct? rigorous evaluation
  of large language models for code generation.
\newblock \emph{Advances in Neural Information Processing Systems}, 36,
  2024{\natexlab{a}}.

\bibitem[Liu et~al.(2024{\natexlab{b}})Liu, Zhou, Hua, Chong, Tian, Liu, Wang,
  You, Guo, Zhu, et~al.]{liu2024cmexam}
Junling Liu, Peilin Zhou, Yining Hua, Dading Chong, Zhongyu Tian, Andrew Liu,
  Helin Wang, Chenyu You, Zhenhua Guo, Lei Zhu, et~al.
\newblock Benchmarking large language models on cmexam-a comprehensive chinese
  medical exam dataset.
\newblock \emph{Advances in Neural Information Processing Systems}, 36,
  2024{\natexlab{b}}.

\bibitem[Loshchilov \& Hutter(2017)Loshchilov and
  Hutter]{loshchilov2017decoupled}
Ilya Loshchilov and Frank Hutter.
\newblock Decoupled weight decay regularization.
\newblock \emph{arXiv preprint arXiv:1711.05101}, 2017.

\bibitem[Maini et~al.(2024)Maini, Seto, Bai, Grangier, Zhang, and
  Jaitly]{maini2024rephrasingwebrecipecompute}
Pratyush Maini, Skyler Seto, He~Bai, David Grangier, Yizhe Zhang, and Navdeep
  Jaitly.
\newblock Rephrasing the web: A recipe for compute and data-efficient language
  modeling, 2024.
\newblock URL \url{https://arxiv.org/abs/2401.16380}.

\bibitem[Men et~al.(2024)Men, Xu, Zhang, Wang, Lin, Lu, Han, and
  Chen]{men2024shortgpt}
Xin Men, Mingyu Xu, Qingyu Zhang, Bingning Wang, Hongyu Lin, Yaojie Lu, Xianpei
  Han, and Weipeng Chen.
\newblock Shortgpt: Layers in large language models are more redundant than you
  expect.
\newblock \emph{arXiv preprint arXiv:2403.03853}, 2024.

\bibitem[OpenAI(2022)]{chatgpt}
OpenAI.
\newblock Introducing chatgpt.
\newblock \emph{Blog post openai.com/blog/chatgpt}, 2022.

\bibitem[OpenAI(2023)]{OpenAI2023GPT4TR}
OpenAI.
\newblock Gpt-4 technical report.
\newblock \emph{ArXiv}, abs/2303.08774, 2023.

\bibitem[Penedo et~al.(2024)Penedo, Kydlíček, allal, Lozhkov, Mitchell,
  Raffel, Werra, and Wolf]{penedo2024finewebdatasetsdecantingweb}
Guilherme Penedo, Hynek Kydlíček, Loubna~Ben allal, Anton Lozhkov, Margaret
  Mitchell, Colin Raffel, Leandro~Von Werra, and Thomas Wolf.
\newblock The fineweb datasets: Decanting the web for the finest text data at
  scale, 2024.
\newblock URL \url{https://arxiv.org/abs/2406.17557}.

\bibitem[Peng et~al.(2023)Peng, Alcaide, Anthony, Albalak, Arcadinho, Biderman,
  Cao, Cheng, Chung, Derczynski, et~al.]{peng2023rwkv}
Bo~Peng, Eric Alcaide, Quentin Anthony, Alon Albalak, Samuel Arcadinho, Stella
  Biderman, Huanqi Cao, Xin Cheng, Michael Chung, Leon Derczynski, et~al.
\newblock Rwkv: Reinventing rnns for the transformer era.
\newblock In \emph{Findings of the Association for Computational Linguistics:
  EMNLP 2023}, pp.\  14048--14077, 2023.

\bibitem[Qwen et~al.(2025)Qwen, :, Yang, Yang, Zhang, Hui, Zheng, Yu, Li, Liu,
  Huang, Wei, Lin, Yang, Tu, Zhang, Yang, Yang, Zhou, Lin, Dang, Lu, Bao, Yang,
  Yu, Li, Xue, Zhang, Zhu, Men, Lin, Li, Tang, Xia, Ren, Ren, Fan, Su, Zhang,
  Wan, Liu, Cui, Zhang, and Qiu]{qwen2025qwen25technicalreport}
Qwen, :, An~Yang, Baosong Yang, Beichen Zhang, Binyuan Hui, Bo~Zheng, Bowen Yu,
  Chengyuan Li, Dayiheng Liu, Fei Huang, Haoran Wei, Huan Lin, Jian Yang,
  Jianhong Tu, Jianwei Zhang, Jianxin Yang, Jiaxi Yang, Jingren Zhou, Junyang
  Lin, Kai Dang, Keming Lu, Keqin Bao, Kexin Yang, Le~Yu, Mei Li, Mingfeng Xue,
  Pei Zhang, Qin Zhu, Rui Men, Runji Lin, Tianhao Li, Tianyi Tang, Tingyu Xia,
  Xingzhang Ren, Xuancheng Ren, Yang Fan, Yang Su, Yichang Zhang, Yu~Wan,
  Yuqiong Liu, Zeyu Cui, Zhenru Zhang, and Zihan Qiu.
\newblock Qwen2.5 technical report, 2025.
\newblock URL \url{https://arxiv.org/abs/2412.15115}.

\bibitem[Romanov \& Shivade(2018)Romanov and Shivade]{romanov2018mednli}
Alexey Romanov and Chaitanya Shivade.
\newblock Lessons from natural language inference in the clinical domain.
\newblock 2018.
\newblock URL \url{http://arxiv.org/abs/1808.06752}.

\bibitem[Sakaguchi et~al.(2021)Sakaguchi, Bras, Bhagavatula, and
  Choi]{sakaguchi2021winogrande}
Keisuke Sakaguchi, Ronan~Le Bras, Chandra Bhagavatula, and Yejin Choi.
\newblock Winogrande: An adversarial winograd schema challenge at scale.
\newblock \emph{Communications of the ACM}, 64\penalty0 (9):\penalty0 99--106,
  2021.

\bibitem[Sanaei et~al.(2023)Sanaei, Ravari, and Abolghasemi]{sanaei2023chatgpt}
Mohammad-Javad Sanaei, Mehrnaz~Sadat Ravari, and Hassan Abolghasemi.
\newblock Chatgpt in medicine: Opportunity and challenges.
\newblock \emph{Iranian Journal of Blood and Cancer}, 15\penalty0 (3):\penalty0
  60--67, 2023.

\bibitem[Schulman et~al.(2017)Schulman, Wolski, Dhariwal, Radford, and
  Klimov]{schulman2017proximal}
John Schulman, Filip Wolski, Prafulla Dhariwal, Alec Radford, and Oleg Klimov.
\newblock Proximal policy optimization algorithms.
\newblock \emph{arXiv preprint arXiv:1707.06347}, 2017.

\bibitem[Sennrich(2015)]{sennrich2015neural}
Rico Sennrich.
\newblock Neural machine translation of rare words with subword units.
\newblock \emph{arXiv preprint arXiv:1508.07909}, 2015.

\bibitem[Shazeer(2020)]{shazeer2020glu}
Noam Shazeer.
\newblock Glu variants improve transformer.
\newblock \emph{arXiv preprint arXiv:2002.05202}, 2020.

\bibitem[Shin et~al.(2020)Shin, Zhang, Bakhturina, Puri, Patwary, Shoeybi, and
  Mani]{shin2020biomegatron}
Hoo-Chang Shin, Yang Zhang, Evelina Bakhturina, Raul Puri, Mostofa Patwary,
  Mohammad Shoeybi, and Raghav Mani.
\newblock Biomegatron: larger biomedical domain language model.
\newblock In \emph{Proceedings of the 2020 Conference on Empirical Methods in
  Natural Language Processing (EMNLP)}, pp.\  4700--4706, 2020.

\bibitem[Singhal et~al.(2025)Singhal, Tu, Gottweis, Sayres, Wulczyn, Amin, Hou,
  Clark, Pfohl, Cole-Lewis, et~al.]{singhal2025toward}
Karan Singhal, Tao Tu, Juraj Gottweis, Rory Sayres, Ellery Wulczyn, Mohamed
  Amin, Le~Hou, Kevin Clark, Stephen~R Pfohl, Heather Cole-Lewis, et~al.
\newblock Toward expert-level medical question answering with large language
  models.
\newblock \emph{Nature Medicine}, pp.\  1--8, 2025.

\bibitem[So et~al.(2021)So, Ma{\'n}ke, Liu, Dai, Shazeer, and
  Le]{so2021searching}
David So, Wojciech Ma{\'n}ke, Hanxiao Liu, Zihang Dai, Noam Shazeer, and Quoc~V
  Le.
\newblock Searching for efficient transformers for language modeling.
\newblock \emph{Advances in neural information processing systems},
  34:\penalty0 6010--6022, 2021.

\bibitem[Spurlock et~al.(2024)Spurlock, Acun, Saka, and
  Nasraoui]{spurlock2024chatgpt}
Kyle~Dylan Spurlock, Cagla Acun, Esin Saka, and Olfa Nasraoui.
\newblock Chatgpt for conversational recommendation: Refining recommendations
  by reprompting with feedback.
\newblock \emph{arXiv preprint arXiv:2401.03605}, 2024.

\bibitem[Su et~al.(2024)Su, Ahmed, Lu, Pan, Bo, and Liu]{su2024roformer}
Jianlin Su, Murtadha Ahmed, Yu~Lu, Shengfeng Pan, Wen Bo, and Yunfeng Liu.
\newblock Roformer: Enhanced transformer with rotary position embedding.
\newblock \emph{Neurocomputing}, 568:\penalty0 127063, 2024.

\bibitem[Tang et~al.(2024)Tang, Ranjan, Pangarkar, Liang, Wang, An, Rao, Jin,
  Wang, Cheng, Sun, Mu, Miller, Ma, Peng, Liu, and Xing]{txt360}
Liping Tang, Nikhil Ranjan, Omkar Pangarkar, Xuezhi Liang, Zhen Wang, Li~An,
  Bhaskar Rao, Linghao Jin, Huijuan Wang, Zhoujun Cheng, Suqi Sun, Cun Mu,
  Victor Miller, Xuezhe Ma, Yue Peng, Zhengzhong Liu, and Eric~P. Xing.
\newblock Txt360: A top-quality llm pre-training dataset requires the perfect
  blend.
\newblock \url{https://huggingface.co/spaces/LLM360/TxT360}, 2024.

\bibitem[Taylor et~al.(2022)Taylor, Kardas, Cucurull, Scialom, Hartshorn,
  Saravia, Poulton, Kerkez, and Stojnic]{taylor2022galactica}
Ross Taylor, Marcin Kardas, Guillem Cucurull, Thomas Scialom, Anthony
  Hartshorn, Elvis Saravia, Andrew Poulton, Viktor Kerkez, and Robert Stojnic.
\newblock Galactica: A large language model for science.
\newblock \emph{arXiv preprint arXiv:2211.09085}, 2022.

\bibitem[Team et~al.(2024)Team, Riviere, Pathak, Sessa, Hardin, Bhupatiraju,
  Hussenot, Mesnard, Shahriari, Ram{\'e}, et~al.]{team2024gemma}
Gemma Team, Morgane Riviere, Shreya Pathak, Pier~Giuseppe Sessa, Cassidy
  Hardin, Surya Bhupatiraju, L{\'e}onard Hussenot, Thomas Mesnard, Bobak
  Shahriari, Alexandre Ram{\'e}, et~al.
\newblock Gemma 2: Improving open language models at a practical size.
\newblock \emph{arXiv preprint arXiv:2408.00118}, 2024.

\bibitem[Touvron et~al.(2023)Touvron, Lavril, Izacard, Martinet, Lachaux,
  Lacroix, Rozière, Goyal, Hambro, Azhar, Rodriguez, Joulin, Grave, and
  Lample]{touvron2023llama}
Hugo Touvron, Thibaut Lavril, Gautier Izacard, Xavier Martinet, Marie-Anne
  Lachaux, Timothée Lacroix, Baptiste Rozière, Naman Goyal, Eric Hambro,
  Faisal Azhar, Aurelien Rodriguez, Armand Joulin, Edouard Grave, and Guillaume
  Lample.
\newblock Llama: Open and efficient foundation language models, 2023.

\bibitem[Vaswani et~al.(2017)Vaswani, Shazeer, Parmar, Uszkoreit, Jones, Gomez,
  Kaiser, and Polosukhin]{vaswani2017attention}
Ashish Vaswani, Noam Shazeer, Niki Parmar, Jakob Uszkoreit, Llion Jones,
  Aidan~N Gomez, {\L}ukasz Kaiser, and Illia Polosukhin.
\newblock Attention is all you need.
\newblock \emph{Advances in neural information processing systems}, 30, 2017.

\bibitem[Waleffe et~al.(2024)Waleffe, Byeon, Riach, Norick, Korthikanti, Dao,
  Gu, Hatamizadeh, Singh, Narayanan, Kulshreshtha, Singh, Casper, Kautz,
  Shoeybi, and Catanzaro]{Waleffe2024AnES}
Roger Waleffe, Wonmin Byeon, Duncan Riach, Brandon Norick, Vijay~Anand
  Korthikanti, Tri Dao, Albert Gu, Ali Hatamizadeh, Sudhakar Singh, Deepak
  Narayanan, Garvit Kulshreshtha, Vartika Singh, Jared Casper, Jan Kautz,
  Mohammad Shoeybi, and Bryan Catanzaro.
\newblock An empirical study of mamba-based language models.
\newblock \emph{ArXiv}, abs/2406.07887, 2024.
\newblock URL \url{https://api.semanticscholar.org/CorpusID:270391285}.

\bibitem[Wang \& Zhang(2024)Wang and Zhang]{wang2024large}
Dandan Wang and Shiqing Zhang.
\newblock Large language models in medical and healthcare fields: applications,
  advances, and challenges.
\newblock \emph{Artificial Intelligence Review}, 57\penalty0 (11):\penalty0
  299, 2024.

\bibitem[Wang et~al.(2024{\natexlab{a}})Wang, Ran, Tang, Chang, Chuang, Liu,
  Braverman, Liu, and Hu]{wang2024redisqa}
Guanchu Wang, Junhao Ran, Ruixiang Tang, Chia-Yuan Chang, Yu-Neng Chuang, Zirui
  Liu, Vladimir Braverman, Zhandong Liu, and Xia Hu.
\newblock Assessing and enhancing large language models in rare disease
  question-answering.
\newblock \emph{arXiv preprint arXiv:2408.08422}, 2024{\natexlab{a}}.

\bibitem[Wang et~al.(2024{\natexlab{b}})Wang, Zhao, Qiang, Li, Liu, Xi, Du,
  Qin, and Liu]{10.1145/3686807}
Haochun Wang, Sendong Zhao, Zewen Qiang, Zijian Li, Chi Liu, Nuwa Xi, Yanrui
  Du, Bing Qin, and Ting Liu.
\newblock Knowledge-tuning large language models with structured medical
  knowledge bases for trustworthy response generation in chinese.
\newblock \emph{ACM Trans. Knowl. Discov. Data}, August 2024{\natexlab{b}}.
\newblock ISSN 1556-4681.
\newblock \doi{10.1145/3686807}.
\newblock URL \url{https://doi.org/10.1145/3686807}.
\newblock Just Accepted.

\bibitem[Wang et~al.(2023)Wang, Chen, Song, Zhang, Chen, Xiao, Jiang, Li, Wan,
  Wang, et~al.]{wang2023cmb}
Xidong Wang, Guiming~Hardy Chen, Dingjie Song, Zhiyi Zhang, Zhihong Chen,
  Qingying Xiao, Feng Jiang, Jianquan Li, Xiang Wan, Benyou Wang, et~al.
\newblock Cmb: A comprehensive medical benchmark in chinese.
\newblock \emph{arXiv preprint arXiv:2308.08833}, 2023.

\bibitem[Wei et~al.(2022)Wei, Wang, Schuurmans, Bosma, Chi, Xia, Le, and
  Zhou]{Wei2022ChainOT}
Jason Wei, Xuezhi Wang, Dale Schuurmans, Maarten Bosma, Ed~H. Chi, F.~Xia, Quoc
  Le, and Denny Zhou.
\newblock Chain of thought prompting elicits reasoning in large language
  models.
\newblock \emph{ArXiv}, abs/2201.11903, 2022.
\newblock URL \url{https://api.semanticscholar.org/CorpusID:246411621}.

\bibitem[Wei et~al.(2023)Wei, Luan, Liu, Dong, and Wang]{wei2023cmath}
Tianwen Wei, Jian Luan, Wei Liu, Shuang Dong, and Bin Wang.
\newblock Cmath: Can your language model pass chinese elementary school math
  test?, 2023.

\bibitem[Wu et~al.(2024)Wu, Wang, Xiao, Peng, and Fu]{wu2024retrieval}
Wenhao Wu, Yizhong Wang, Guangxuan Xiao, Hao Peng, and Yao Fu.
\newblock Retrieval head mechanistically explains long-context factuality.
\newblock \emph{arXiv preprint arXiv:2404.15574}, 2024.

\bibitem[Xu et~al.(2019)Xu, Sun, Zhang, Zhao, and Lin]{xu2019understanding}
Jingjing Xu, Xu~Sun, Zhiyuan Zhang, Guangxiang Zhao, and Junyang Lin.
\newblock Understanding and improving layer normalization.
\newblock \emph{Advances in neural information processing systems}, 32, 2019.

\bibitem[Xu et~al.(2024{\natexlab{a}})Xu, Cheng, Wang, and Chen]{xu2024kv}
Mingyu Xu, Wei Cheng, Bingning Wang, and Weipeng Chen.
\newblock Kv shifting attention enhances language modeling.
\newblock \emph{arXiv preprint arXiv:2411.19574}, 2024{\natexlab{a}}.

\bibitem[Xu et~al.(2024{\natexlab{b}})Xu, Men, Wang, Zhang, Lin, Han,
  et~al.]{xu2024base}
Mingyu Xu, Xin Men, Bingning Wang, Qingyu Zhang, Hongyu Lin, Xianpei Han,
  et~al.
\newblock Base of rope bounds context length.
\newblock In \emph{The Thirty-eighth Annual Conference on Neural Information
  Processing Systems}, 2024{\natexlab{b}}.

\bibitem[Yan et~al.(2024)Yan, Liu, Wu, Chen, Wang, Chai, Wang, Zhao, Zhang,
  Zhang, et~al.]{yan2024clinicalbench}
Weixiang Yan, Haitian Liu, Tengxiao Wu, Qian Chen, Wen Wang, Haoyuan Chai,
  Jiayi Wang, Weishan Zhao, Yixin Zhang, Renjun Zhang, et~al.
\newblock Clinicallab: Aligning agents for multi-departmental clinical
  diagnostics in the real world.
\newblock \emph{arXiv preprint arXiv:2406.13890}, 2024.

\bibitem[Yang et~al.(2023)Yang, Xiao, Wang, Zhang, Bian, Yin, Lv, Pan, Wang,
  Yan, et~al.]{yang2023baichuan}
Aiyuan Yang, Bin Xiao, Bingning Wang, Borong Zhang, Ce~Bian, Chao Yin, Chenxu
  Lv, Da~Pan, Dian Wang, Dong Yan, et~al.
\newblock Baichuan 2: Open large-scale language models.
\newblock \emph{arXiv preprint arXiv:2309.10305}, 2023.

\bibitem[Yang et~al.(2025{\natexlab{a}})Yang, Venkitesh, Talupuru, Lin, Cairuz,
  Blunsom, and Locatelli]{yang2025rope}
Bowen Yang, Bharat Venkitesh, Dwarak Talupuru, Hangyu Lin, David Cairuz, Phil
  Blunsom, and Acyr Locatelli.
\newblock Rope to nope and back again: A new hybrid attention strategy.
\newblock \emph{arXiv preprint arXiv:2501.18795}, 2025{\natexlab{a}}.

\bibitem[Yang et~al.(2024)Yang, Wang, Zhang, Shen, and
  Kim]{yang2024parallelizing}
Songlin Yang, Bailin Wang, Yu~Zhang, Yikang Shen, and Yoon Kim.
\newblock Parallelizing linear transformers with the delta rule over sequence
  length.
\newblock \emph{arXiv preprint arXiv:2406.06484}, 2024.

\bibitem[Yang et~al.(2025{\natexlab{b}})Yang, Kautz, and
  Hatamizadeh]{yang2025gated}
Songlin Yang, Jan Kautz, and Ali Hatamizadeh.
\newblock Gated delta networks: Improving mamba2 with delta rule.
\newblock In \emph{The Thirteenth International Conference on Learning
  Representations}, 2025{\natexlab{b}}.
\newblock URL \url{https://openreview.net/forum?id=r8H7xhYPwz}.

\bibitem[Zellers et~al.(2019)Zellers, Holtzman, Bisk, Farhadi, and
  Choi]{zellers2019hellaswag}
Rowan Zellers, Ari Holtzman, Yonatan Bisk, Ali Farhadi, and Yejin Choi.
\newblock Hellaswag: Can a machine really finish your sentence?
\newblock \emph{arXiv preprint arXiv:1905.07830}, 2019.

\bibitem[Zeng et~al.(2024)Zeng, Liu, Ma, Yang, Zhang, and Wang]{zeng2024token}
Yongcheng Zeng, Guoqing Liu, Weiyu Ma, Ning Yang, Haifeng Zhang, and Jun Wang.
\newblock Token-level direct preference optimization.
\newblock \emph{arXiv preprint arXiv:2404.11999}, 2024.

\bibitem[Zhang \& Sennrich(2019)Zhang and Sennrich]{zhang2019root}
Biao Zhang and Rico Sennrich.
\newblock Root mean square layer normalization.
\newblock \emph{Advances in Neural Information Processing Systems}, 32, 2019.

\bibitem[Zhang et~al.(2023)Zhang, Chen, Jiang, Yu, Chen, Li, Chen, Wu, Zhang,
  Xiao, et~al.]{zhang2023huatuogpt}
Hongbo Zhang, Junying Chen, Feng Jiang, Fei Yu, Zhihong Chen, Jianquan Li,
  Guiming Chen, Xiangbo Wu, Zhiyi Zhang, Qingying Xiao, et~al.
\newblock Huatuogpt, towards taming language model to be a doctor.
\newblock \emph{arXiv preprint arXiv:2305.15075}, 2023.

\bibitem[Zhang et~al.(2024)Zhang, Li, Zong, Ying, He, and
  Qiu]{zhang2024evaluatingperformancelargelanguage}
Xiaotian Zhang, Chunyang Li, Yi~Zong, Zhengyu Ying, Liang He, and Xipeng Qiu.
\newblock Evaluating the performance of large language models on gaokao
  benchmark, 2024.
\newblock URL \url{https://arxiv.org/abs/2305.12474}.

\bibitem[Zhang \& Yang(2023)Zhang and Yang]{zhang2023xuanyuan}
Xuanyu Zhang and Qing Yang.
\newblock Xuanyuan 2.0: A large chinese financial chat model with hundreds of
  billions parameters.
\newblock In \emph{Proceedings of the 32nd ACM international conference on
  information and knowledge management}, pp.\  4435--4439, 2023.

\bibitem[Zhao et~al.(2023)Zhao, Jin, Chen, Peng, and Yu]{zhao2023large}
Zhengyun Zhao, Qiao Jin, Fangyuan Chen, Tuorui Peng, and Sheng Yu.
\newblock A large-scale dataset of patient summaries for retrieval-based
  clinical decision support systems.
\newblock \emph{Scientific data}, 10\penalty0 (1):\penalty0 909, 2023.

\bibitem[Zhong et~al.(2019)Zhong, Xiao, Tu, Zhang, Liu, and
  Sun]{zhong2019jecqalegaldomainquestionanswering}
Haoxi Zhong, Chaojun Xiao, Cunchao Tu, Tianyang Zhang, Zhiyuan Liu, and Maosong
  Sun.
\newblock Jec-qa: A legal-domain question answering dataset, 2019.
\newblock URL \url{https://arxiv.org/abs/1911.12011}.

\bibitem[Zhuo et~al.(2024)Zhuo, Vu, Chim, Hu, Yu, Widyasari, Yusuf, Zhan, He,
  Paul, et~al.]{zhuo2024bigcodebench}
Terry~Yue Zhuo, Minh~Chien Vu, Jenny Chim, Han Hu, Wenhao Yu, Ratnadira
  Widyasari, Imam Nur~Bani Yusuf, Haolan Zhan, Junda He, Indraneil Paul, et~al.
\newblock Bigcodebench: Benchmarking code generation with diverse function
  calls and complex instructions.
\newblock \emph{arXiv preprint arXiv:2406.15877}, 2024.

\end{thebibliography}
\bibliographystyle{baichuan}

\newpage
\appendix
\section{Contributor}\label{sec.contributor}

\subsection{Corresponding Author}
\begin{multicols}{2} %
	Bingning Wang, daniel@baichuan-inc.com
\end{multicols}
\subsection{Core Contributors}
\begin{multicols}{2} %
	Bingning Wang\\  
	Haizhou Zhao\\ 
	Huozhi Zhou\\ 
	Liang Song\\ 
	Mingyu Xu\\ 
	Wei Cheng\\ 
	Xiangrong Zeng\\ 
	Yupeng Zhang\\ 
	Yuqi Huo\\ 
	Zecheng Wang\\ 
	Zhengyun Zhao
\end{multicols}
\subsection{Contributors}
\begin{multicols}{2} %
	Da Pan\\
	Fei Kou\\ 
	Fei Li\\ 
	Fuzhong Chen\\ 
	Guosheng Dong\\ 
	Han Liu\\ 
	Hongda Zhang\\ 
	Jin He \\
	Jinjie Yang\\ 
	Kangxi Wu\\ 
	Kegeng Wu\\ 
	Lei Su\\ 
	Linlin Niu\\ 
	Linzhuang Sun\\ 
	Mang Wang\\ 
	Pengcheng Fan\\ 
	Qianli Shen\\
	Rihui Xin\\ 
	Shunya Dang\\ 
	Songchi Zhou\\ 
	Weipeng Chen\\ 
	Wenjing Luo\\ 
	Xin Chen\\ 
	Xin Men\\ 
	Xionghai Lin\\ 
	Xuezhen Dong\\ 
	Yan Zhang\\ 
	Yifei Duan\\ 
	Yuyan Zhou\\ 
	Zhi Ma\\ 
	Zhiying Wu
\end{multicols}
\section{Examples of synthetic data}
We present several examples of our synthetic data in this section.
\subsection{Example 1}
\paragraph{Question}
Which surgical procedure is most appropriate for treating an aneurysm in the lower part of the thoracic aorta using an invasive method?

A. Endovascular aneurysm repair (EVAR)

B. Thoracic endovascular aortic repair (TEVAR)

C. Open surgical repair of descending thoracic aortic aneurysm

D. Hybrid aortic arch repair

E. Aortic valve replacement with ascending aorta graft

\paragraph{Answer}
Let's analyze each option to determine which surgical procedure is the most appropriate for treating an aneurysm in the lower part of the thoracic aorta using an invasive method.

A. Endovascular aneurysm repair (EVAR)
Endovascular aneurysm repair (EVAR) is a minimally invasive procedure that involves inserting a stent graft through the blood vessels to repair the aneurysm. However, EVAR is typically used for abdominal aortic aneurysms rather than thoracic aortic aneurysms. Therefore, this option is not the most appropriate for the given condition.

B. Thoracic endovascular aortic repair (TEVAR)
Thoracic endovascular aortic repair (TEVAR) is a minimally invasive procedure used to treat thoracic aortic aneurysms. It involves the placement of a stent graft inside the thoracic aorta to reinforce the aneurysm site. While TEVAR is used for thoracic aortic aneurysms, it is considered less invasive compared to open surgical methods, so it does not meet the criteria of being an invasive method.

C. Open surgical repair of descending thoracic aortic aneurysm
Open surgical repair of a descending thoracic aortic aneurysm involves a direct surgical approach to repair the aneurysm by opening the chest cavity. This is a traditional open surgery method and is considered invasive. This option matches the requirement of treating an aneurysm in the lower part of the thoracic aorta using an invasive method.

D. Hybrid aortic arch repair
Hybrid aortic arch repair combines open surgical techniques with endovascular methods to treat complex aortic arch aneurysms. While this procedure involves both invasive and minimally invasive components, it is primarily used for aortic arch aneurysms and not specifically for descending thoracic aortic aneurysms.

E. Aortic valve replacement with ascending aorta graft
Aortic valve replacement with ascending aorta graft involves replacing the aortic valve and repairing the ascending aorta. This procedure is not directly related to the descending thoracic aorta and is therefore not appropriate for the given condition.

Based on the analysis, the most appropriate surgical procedure for treating an aneurysm in the lower part of the thoracic aorta using an invasive method is: C. Open surgical repair of descending thoracic aortic aneurysm.
\subsection{Example 2}
\paragraph{Question}
A 24-year-old female patient started to experience involuntary tremors in both thumbs half a year ago. The tremors gradually progressed to involve both hands and were more obvious during movement. The symptoms gradually worsened, accompanied by tremors in both lower limbs and unsteady walking. She gradually developed mild speech difficulties, slurred speech, choking during swallowing, as well as personality changes, with obvious mood swings and irritability. Laboratory tests indicated abnormal liver function, abnormal kidney function, low white blood cell count, and a significantly decreased ceruloplasmin. Head MRI showed multiple abnormal signal changes in the bilateral basal ganglia, thalamus, midbrain, and pons, along with demyelination in the white matter of the left frontal lobe. Regarding the pathogenesis of the patient's neurological symptoms in this case, which of the following most accurately describes its pathological process?

A. Immune - mediated central nervous system demyelination

B. Abnormal accumulation of copper ions in the basal ganglia and other brain regions, generating a large number of free radicals leading to neuronal damage

C. Gene mutations cause abnormal myelin proteins, resulting in widespread central nervous system demyelination

D. Cerebral microvascular lesions lead to repeated small - area infarctions, causing nerve function impairment

E. Deposition of iron ions in the brain, triggering oxidative stress and nerve damage
\paragraph{Answer}

To determine the correct pathological process in this patient's case, we need to analyze each option in detail, considering the patient's symptoms, laboratory results, and imaging findings:  

A. **Immune-mediated central nervous system demyelination**:  
- This option refers to conditions such as multiple sclerosis, where the immune system attacks the myelin sheath of neurons in the central nervous system, leading to demyelination.  
- Symptoms typically include a range of neurological deficits, but the specific combination of liver and kidney dysfunction, decreased ceruloplasmin, and the MRI findings of abnormal signal changes in the basal ganglia and thalamus are not characteristic of multiple sclerosis or similar demyelinating diseases.  

B. **Abnormal accumulation of copper ions in the basal ganglia and other brain regions, generating a large number of free radicals leading to neuronal damage**:  
- This describes Wilson's disease, a genetic disorder causing copper accumulation in the liver, brain, and other tissues. This can lead to hepatic and neurological symptoms, including tremors, personality changes, speech difficulties, and liver dysfunction.  
- The patient's abnormal liver and kidney function, low ceruloplasmin, and MRI findings involving the basal ganglia support this diagnosis.  

C. **Gene mutations cause abnormal myelin proteins, resulting in widespread central nervous system demyelination**:  
- This option refers to leukodystrophies, which are genetic disorders affecting the white matter of the brain due to defects in myelin production or maintenance.  
- While demyelination was noted in the white matter of the left frontal lobe, the broad range of systemic symptoms and specific MRI findings are not typical for leukodystrophies.  

D. **Cerebral microvascular lesions lead to repeated small-area infarctions, causing nerve function impairment**:  
- This option is characteristic of conditions like small vessel disease or cerebral autosomal dominant arteriopathy with subcortical infarcts and leukoencephalopathy (CADASIL).  
- However, the systemic symptoms including liver and kidney dysfunction, low ceruloplasmin, and specific areas of brain involvement (basal ganglia, thalamus) are not typical of these conditions.  

E. **Deposition of iron ions in the brain, triggering oxidative stress and nerve damage**:  
- This describes conditions such as neurodegeneration with brain iron accumulation (NBIA), which can cause movement disorders and cognitive impairment.  
- While some neurological symptoms might overlap, the systemic findings (especially liver dysfunction and low ceruloplasmin) are not explained by this condition.   

**Summary:**  
The most accurate description of the pathological process in this patient's case is **B. Abnormal accumulation of copper ions in the basal ganglia and other brain regions, generating a large number of free radicals leading to neuronal damage**. This aligns with Wilson's disease, given the combination of neurological symptoms, systemic findings, and imaging results.
\subsection{Example 3}
\paragraph{Question}
What role does vitamin K play in blood coagulation? What specific health problems can be caused by vitamin K deficiency?
\paragraph{Answer}
Vitamin K plays a crucial role in blood coagulation. It is essential for the synthesis of multiple clotting factors (such as clotting factors II (prothrombin), VII, IX, and X) and anticoagulant proteins (such as proteins C and S). These clotting factors are produced in the liver and participate in the blood coagulation process, helping to stop bleeding.  

Specifically, vitamin K is involved in the r-carboxylation of these clotting factors, a process that enables these proteins to bind to calcium ions, thereby functioning effectively in the blood coagulation process. Without sufficient vitamin K, the activity of these clotting factors is significantly reduced, leading to impaired blood clotting.  

Deficiency in vitamin K can lead to the following specific health problems:  

1. Bleeding tendency: Due to impaired generation and function of clotting factors, patients with vitamin K deficiency are prone to bleeding issues. This may manifest as skin bruising, nosebleeds, gum bleeding, gastrointestinal bleeding, and even more severe internal bleeding.  

2. Hemorrhagic disease of the newborn: Newborns are particularly susceptible to vitamin K deficiency because their intestines have not yet established enough bacterial colonies to synthesize vitamin K, and breast milk contains relatively low amounts of vitamin K. Therefore, newborns who have not received vitamin K injections may experience severe bleeding problems, such as intracranial hemorrhage.  

3. Bone health issues: Vitamin K not only plays a role in blood coagulation but also in bone health. It is involved in the carboxylation of osteocalcin, a protein that regulates bone mineralization, which is crucial for bone health and strength. Vitamin K deficiency can lead to decreased bone density and an increased risk of fractures.  

4. Cardiovascular health problems: Vitamin K also plays a role in regulating vascular calcification. A deficiency in vitamin K may lead to increased vascular calcification, thereby raising the risk of cardiovascular diseases.  

Measures to prevent vitamin K deficiency include consuming foods rich in vitamin K (such as leafy green vegetables, fermented foods, and certain plant oils) and using vitamin K supplements when necessary. In newborns, vitamin K is typically administered via injection shortly after birth to prevent hemorrhagic disease. For patients at risk of coagulation disorders, especially those taking anticoagulant medications, vitamin K levels should be monitored and managed under a doctor's guidance.
\subsection{Example 4}
\paragraph{Question}
In a study, the researchers hoped to compare the effects of two treatments on the coronary sinus blood flow in animals. It was known that the average blood flow increase of treatment A was 1.8 ml/min, and that of treatment B was 2.4 ml/min, and the standard deviation of both treatments was 1.0 ml/min. If the test level $\alpha$ was set at 0.05 and the type II error probability $\beta$ was 0.10, how many experimental animals would the researchers need to draw the conclusion that there was a statistically significant difference between the two treatments?
\paragraph{Answer}
To calculate the sample size needed to detect a statistically significant difference between the two treatments, we can use the formula for comparing two means in a two-sample t-test.

The formula for calculating the sample size n in a two-sample t-test is:

$$n = \frac{(z_{1-\alpha/2}+z_{1 - \beta})^2\times2\sigma^2}{(\mu_1-\mu_2)^2}$$

where:

- $z_{1-\alpha/2}$ is the critical value corresponding to the significance level $\alpha$. For a two-tailed test with $\alpha = 0.05$, $z_{1-\alpha/2}=z_{0.975}\approx 1.96$.

- $z_{1-\beta}$ is the critical value corresponding to the power of the test $(1 - \beta)$. Given $\beta=0.10$, then $z_{1-\beta}=z_{0.90}\approx1.28$.

- $\sigma$ is the standard deviation of the two groups, assumed to be equal. Here $\sigma = 1.0$ ml/min.

- $\mu_1$ and $\mu_2$ are the means of the two groups. Here $\mu_1 = 1.8$ ml/min and $\mu_2=2.4$ ml/min, so $\mu_1-\mu_2=- 0.6$ ml/min.

We take the absolute value of $\mu_1-\mu_2$, so $\vert\mu_1-\mu_2\vert=0.6$ ml/min

Substitute the values into the formula:

\[n=\frac{(1.96 + 1.28)^2\times2\times1^2}{(0.6)^2}\]

\[n=\frac{(3.24)^2\times2\times1}{0.36}\]

\[n=\frac{10.4976\times2}{0.36}=\frac{20.9952}{0.36}\approx58.32\]

Since we cannot have a fraction of an animal, we round up to $n = 59$

So the researchers would need $59$ experimental animals in each group to draw the conclusion that there was a statistically significant difference between the two treatments at the given significance level and power of the test. In total, $2\times59=118$ animals are needed for the whole study.

\section{Tokenizer Details}
\label{tokenizer_efficiency}

\begin{figure}[ht]
	\centering
	\includegraphics[width=1\textwidth]{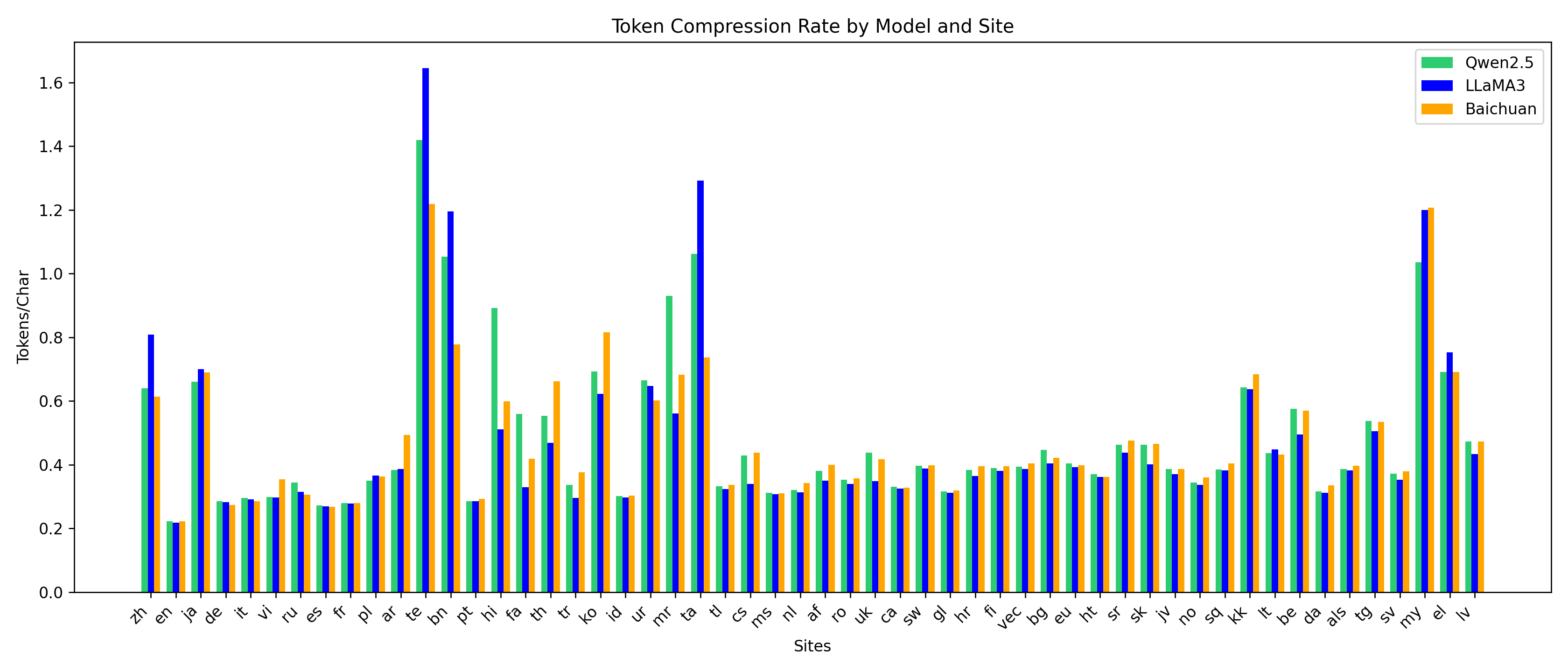}
	\caption{The tokenization efficiency for different models on different languages, the less the better.}
	\label{fig:tokenizer.mutli}
\end{figure}

For the base general tokenizer, we utilize the sentencepiece tokenizer \cite{kudo2018sentencepiece}. For the additional multi-lingual and medical tokenizer, we adopt the huggingface tokenizer\footnote{\url{https://huggingface.co/docs/transformers/en/main_classes/tokenizer}} to train a BPE tokenizer \cite{sennrich2015neural}. And then merge the tokenizer together. The specific rules we used are:
\begin{itemize}
	\item We do not normalized the input tokens since, i.e., the $\mathtt{normalization\_rule\_name}$ is set to $\mathtt{identity}$.
	\item We kept the pieces that only contains whitespace, allowing for more efficient code encoding.
	\item We split numbers into individual digits to better encode numeric data.
	\item The character coverage is set to 0.9999 with rare character falling back to UTF-8 bytes.
\end{itemize}
The tokenization efficiency of our model in different language is shown in Figure \ref{fig:tokenizer.mutli}.

\section{Evaluation Prompts and Examples}
\label{evalpromptsappendix}

\begin{table}[]
	\begin{CJK}{UTF8}{gbsn}
		\begin{tabular}{|l|l|}
			\hline
			\textbf{Benchmark} & \multicolumn{1}{c|}{\textbf{Prompt}}                                                                                              \\ \hline
			MedCalc &
			\makecell[l]{\\You are a medical professional and you have been asked to answer a \\question based on the patient note provided.
				The patient note is a \\record of a patient's visit to the doctor. 
				Note that you should not \\provide a diagnosis, but rather only answer the question based on \\the information provided in the note.
				Your response should be a \\numerical value, which is the answer to the question.\\
				Let's think step by step, and at the end of your response, repeat your\\answer in a new line as the format requested in the question, with \\prefix "The final answer is: ".
			} \\ \hline
			Multiple-Choice    & \makecell[l]{\\Please provide an analysis and the correct answer of the given \\multiple-choice question. Please note that only one answer is correct. }\\ \hline
			\makecell[l]{ClinicalBench--\\Department}  & \makecell[l]{\\您是一位经验丰富的医院导诊员，根据患者提供的性别、年龄和症状\\等信息，请从医院科室列表中选择一个患者最需要前往治疗的科室。\\
				请确保您的回答只包含选中的一个科室，不要包含多余的内容。
				以下\\是医院科室列表：...
			}                           \\ \hline
			\makecell[l]{ClinicalBench--\\Diagnosis} & \makecell[l]{\\您是一位经验丰富的{科室}临床医生，根据给定的病例摘要，请分析并\\给出该病人的主要诊断，即一种对患者身体健康危害最大、最需要治\\疗的疾病的名称。以下是给定的病例摘要：}\\ \hline
			\makecell[l]{ClinicalBench--\\Treatment} & \makecell[l]{\\您是一位经验丰富的{科室}临床医生，根据给定的病例摘要，请分析并\\提供一个专业、详细、全面的治疗计划，请分条列出治疗计划。以下\\是给定的病例摘要：}\\ \hline
			MedNLI &
			\makecell[l]{\\Sentence1: \{\}\\ \\ Sentence2: \{\} \\ \\ Whether sentence 2 can be inferred from sentence 1? Pick a choice \\ from 3 options: entailment, contradiction, or neutral.\\ Only output the option. } \\ \hline
			NEJMQA &
			\makecell[l]{\\Given the patient note, enumerate the top 5 most likely diagnoses.}\\ \hline
			RareArena          & \makecell[l]{\\Given the patient note, enumerate the top 5 most likely diagnoses. \\
				Only consider rare diseases.}  \\ \hline
			RareBench          & \makecell[l]{\\Given the patient symptoms, enumerate the top 5 most likely\\ diagnoses. Only consider rare diseases.}  \\ \hline
		\end{tabular}
	\end{CJK}
	\caption{Prompts used during the response generation.}
	\label{responseprompt}
\end{table}

\begin{table}[]
	\begin{CJK}{UTF8}{gbsn}
		\begin{tabular}{|l|l|}
			\hline
			\textbf{Benchmark} & \multicolumn{1}{c|}{\textbf{Prompt}}                                                                                                  \\ \hline
			Multiple-Choice    & \begin{tabular}[c]{@{}l@{}}\\以下是某个选择题的回答，请抽取其选中的选项标签，以json格式\\返回结果，除此以外不要返回其它任何信息。如：\\ \{\{"select": "A"\}\}\\ \\ 回答：\\ \{\}\end{tabular} \\ \hline
			CMBClin &
			\begin{tabular}[c]{@{}l@{}}\\请参考正确答案的内容，给候选答案进行评分。\\ 1、分数为整数，取值范围为0、1、2、3或者4分。\\ 2、评分最低分为0分，表示候选答案与正确答案毫不相关；\\ 3、评分最高分为4分，表示候选答案涵盖了所有要点；\\ 4、如果候选答案相比正确答案缺少了部分要点，则要酌情扣分；\\ 5、以json格式返回分数以及理由，除此以外不要返回其它任何\\信息。如：\\ \{\{"score": 4, "reason": "xxx"\}\}\\ \\ 正确答案：\\ \{\}\\ \\ 候选答案：\\ \{\}\end{tabular} \\ \hline
			\makecell[l]{NEJMQA\\RareArena\\RareBench} &
			\begin{tabular}[c]{@{}l@{}}\\1、在一个诊断任务中，正确的诊断是\{\}，以下是学生给出的五个鉴\\别诊断，请判断正确诊断是否被包含其中（包括学生回答中包含了\\正确诊断上级概念的情况，即学生未明确判断亚型也算正确）。\\ 2、如果包含了正确诊断，则分数为：1，否则分数为：0。\\ 3、以json格式返回结果及理由，除此以外不要返回其它任何信息。\\如：\\ \{\{"score": 1, "reason": "xxx"\}\}\\ \\ 学生回答：\\ \{\}\end{tabular} \\ \hline
			MedCalc &
			\begin{tabular}[c]{@{}l@{}}\\请判断给定的回答与正确答案是否一致。如果一致，得分为：1，\\否则为：0。 \\具体来说，正确答案可能有以下几种情况：\\ 1. 单一数字（整数或浮点数），直接判断回答是否完全正确。\\ 2. 日期，格式可能是mm/dd/yyyy，或者(4 weeks, 3 days)，直接\\判断回答是否完整正确（忽略格式差异）。\\ 3. 两个数字，代表正确的范围（下界，上界），请判断回答是否处\\于这个范围内。 以json格式返回结果及理由，除此以外不\\要返回其它任何信息。如：\\ \{\{"score": 1, "reason": "xxx"\}\}\\ \\ 正确答案：\\ \{\}\\ \\ 回答：\\ \{\}\end{tabular} \\ \hline
		\end{tabular}
	\end{CJK}
	\caption{Scoring prompts of different benchmarks.}
	\label{scoreprompt}
\end{table}

\begin{CJK}{UTF8}{gbsn}
	\begin{table}[]
		\begin{tabular}{|l |l |}
			\hline
			\textbf{Question} &
			\textbf{Response} \\ \hline
			\begin{tabular}[c]{@{}l@{}}患者：被沾有病人血液的实心针头扎了，针头是一天前沾\\
				的病人血液，还有传染疾病的可能吗（女, 年龄26岁）\\ 
				医生：请问您是否知道那位病人的具体健康状况或是否患有\\
				任何传染性疾病？例如，乙肝、丙肝或艾滋病等？\\ 
				患者：不知道。\\ 
				医生：请问您是否接种过乙型肝炎疫苗？\\ 
				患者：接种过。\\ 
				医生：请问您被扎到的具体部位是哪里？伤口的深度如何？\\
				是否有出血？\\ 
				患者：大拇指根部，深度大约0.5厘米，出血了。\\ 
				根据以上信息，若该病人同时患有乙肝、丙肝或艾滋病，\\患者最可能感染哪种疾病，并说明理由。\end{tabular} &
			\begin{tabular}[c]{@{}l@{}}
				丙肝。针头是一天\\
				前沾的病人血液，\\
				HIV在干燥环境中迅\\
				速失去活性，因此一\\
				天后针头上的HIV传\\
				染性极低。患者接种\\
				过乙肝疫苗，对患者\\
				有一定的保护。丙肝\\
				病毒在体外也有一定\\
				的存活时间，因此感\\
				染可能性较大。 
			\end{tabular} \\ 
			
			\hline
			
			\begin{tabular}[c]{@{}l@{}}
				一名7岁女孩被送到急诊科，她有10天的发热和咽喉痛病史，\\
				过去2天症状急剧恶化。她的咽喉痛在右侧更严重。由于疼痛，\\
				她吞咽和张口困难。她还注意到在过去一天中，她的声音\\
				质量发生了变化。她的体温为38.2°C（100.8°F），\\
				脉搏为86次/分钟，呼吸为18次/分钟，血压为110/75 mm Hg。\\
				检查显示颈部淋巴结肿大。口咽检查显示扁桃体发红，\\
				右侧扁桃体柱肿胀。悬雍垂偏向左侧。实验室检查结果显示：\\
				血红蛋白 13.0 g/dL，红细胞压积 39\%，白细胞计数\\
				12,000/mm3，血小板计数 200,000/mm3，\\
				血清钠 138 mEq/L，氯 100 mEq/L，钾 4.5 mEq/L，\\
				碳酸氢盐 24 mEq/L，尿素氮 14.0 mg/dL，\\
				肌酐 1.1 mg/dL。基于这些信息，\\
				请分析最合适的下一步手术处理措施。只给一个答案。 
			\end{tabular} &
			
			\begin{tabular}[c]{@{}l@{}}
				手术处理措施切开\\
				引流
			\end{tabular}\\ 
			
			\hline
			
			\begin{tabular}[c]{@{}l@{}}
				一位78岁女性患者，既往诊断为二尖瓣脱垂，来急诊就诊，\\
				主诉24小时内出现4次鲜红色便血。体格检查显示脉率\\
				120次/分，血压88/58mmHg。应该采取的最佳初步处\\
				理措施是什么？只给一个答案。\end{tabular} &
			
			\begin{tabular}[c]{@{}l@{}}
				立即进行静脉补液\\
				（也即液体复苏） \end{tabular}\\ \hline
			
			\begin{tabular}[c]{@{}l@{}}\#一般资料\#:性别: 男年龄: 43岁职业: 无\\ 
				\#主诉\#:全身黄染、腹胀、尿少、乏力、纳差伴双下肢浮肿半月。\\
				\#现病史\#:患者半月前出现全身黄染，伴随腹胀、尿少、\\
				纳差，曾在外院接受治疗，治疗内容包括抗病毒、抗感染、\\
				补充白蛋白等，但症状未缓解，反而腹胀加重，出现双下肢浮肿，\\
				因此再次入院。患者自述病后精神、食欲减退，睡眠一般，\\ 
				小便颜色深黄，量少，大便干结，无黑便。\\ 
				\#既往史\#:患者既往乙型肝炎病史，无冠心病、高血压、\\
				糖尿病等病史，否认痢疾、伤寒、疟疾、结核病史及密切接\\
				触史。无外伤手术史，无输血史。无药物及食物过敏史，预\\
				防接种史不详。\\
				\#查体\#:中年男性, ......, 神经系统检查未见异常。\\ 
				\#辅助检查\#:- 彩色B超: - 肝实质回声增粗，考虑肝硬化可能 \\
				- 胆囊壁增厚 - 副脾 - 双肾结石 - 腹水- DR胸片: - 心肺膈未\\
				见异常- 血液检查: - 凝血酶原时间: \\
				36.6秒↑ - 肌钙蛋白I: 0.02ng/ml↑ - 丙氨酸氨基转移酶: 71.36U/L↑ \\
				- 天门冬氨酸氨基转移酶: 91.65U/L↑ - 白蛋白: 40.78g/L \\
				- 总胆红素: 689.03μmol/L↑ - 直接胆红素: 391.85μmol/L↑ \\
				- 中性粒细胞比率: 62.5\%
				\\ 根据以上资料，该患者最有可能的诊断及诊断依据是什么？\end{tabular} &
			\begin{tabular}[c]{@{}l@{}}
				\#初步诊断\#:1. . 乙型\\
				肝炎肝硬化失代偿期\\ 
				\#诊断依据\#:1. 患者有\\
				全身黄染、腹胀、尿\\
				少、乏力、纳差伴双\\
				下肢浮肿半月的症状。\\
				2. 既往乙型肝炎病史。\\
				3.查体发现全身皮肤\\
				黏膜重度黄染，巩膜\\
				黄染，腹部膨隆，全\\
				腹轻压痛，腹水征阳\\
				性，双下肢中度浮肿。\\
				4. 辅助检查结果提示\\
				肝硬化，胆囊壁增厚，\\
				双肾结石。\end{tabular} \\ \hline
		\end{tabular}
		\caption{Examples of Baichuan-M1-14B-Instruction.}
		\label{examplesofbc}
	\end{table}
\end{CJK}

\end{document}